\documentclass[conference]{IEEEtran}

\IEEEoverridecommandlockouts 

\usepackage{hyperref}
\hypersetup{
    colorlinks=true,
    linkcolor=blue,
    filecolor=blue,      
    urlcolor=blue,
    citecolor=blue,
    pdftitle={combots},
    pdfpagemode=FullScreen,
    }
\usepackage[T1]{fontenc}
\usepackage{graphicx}
\usepackage{url}
\usepackage[numbers,sort&compress]{natbib}

\usepackage{colortbl} 
\usepackage[dvipsnames]{xcolor}
\reversemarginpar

\newcommand{\figpath}{figures} 

\begin{document}

\title{Design and Fabrication of Soft Locomotion Robots based on Spatial Compliant Mechanisms

}

\author{\IEEEauthorblockN{Andrija Milojevic}
\IEEEauthorblockA{\textit{Department of Mechanical
Engineering} \\ \textit{LUT University} \\ \textit{Lappeenranta, Finland (former)} \\ 
\tt andrijamilojevic87@gmail.com}
\and

\IEEEauthorblockN{Kyrre Glette}
\IEEEauthorblockA{\textit{RITMO, Department of Informatics} \\
\textit{University of Oslo}\\
Oslo, Norway \\
\tt kyrrehg@ifi.uio.no}
}
\maketitle



\begin{abstract}
Soft robotics has emerged as a promising technology that holds great potential for various application areas. This is due to soft materials unique properties, including flexibility, safety, and shock absorption, among others. Despite many advancement in the field, the development of effective design methodologies and production techniques for soft robots remains a challenge. Although numerous robot prototypes have been proposed in recent years, their designs are often complex and difficult to produce. As such, there is a need for more efficient and unified design approaches that can facilitate the production of soft robots with desirable properties. In this paper, we propose a method for designing soft robots using elastic beams and spatial compliant mechanisms. The method is based on an evolutionary approach that enables the creation of designs with both high motion and force transmission ratios. Specifically, we focus on the development of locomotion mechanisms using a central linear actuator. Our approach involves the use of commonly available plastic materials and a 3D printer to manufacture the designs. We demonstrate the feasibility of our approach by presenting experimental results that show successful production and real-world operation. 
Overall, our findings suggest that the use of elastic beams and an evolutionary 
 approach can facilitate the creation of soft robots with desirable locomotion properties, including fast locomotion up to 3.7 body lengths per second, locomotion with a payload, and underwater locomotion.
This method has the potential to enable the development of more efficient and practical soft robots for various applications.

\end{abstract}
  

\section{Introduction}

\begin{figure}
    \centering
    \includegraphics[width=0.48\textwidth]{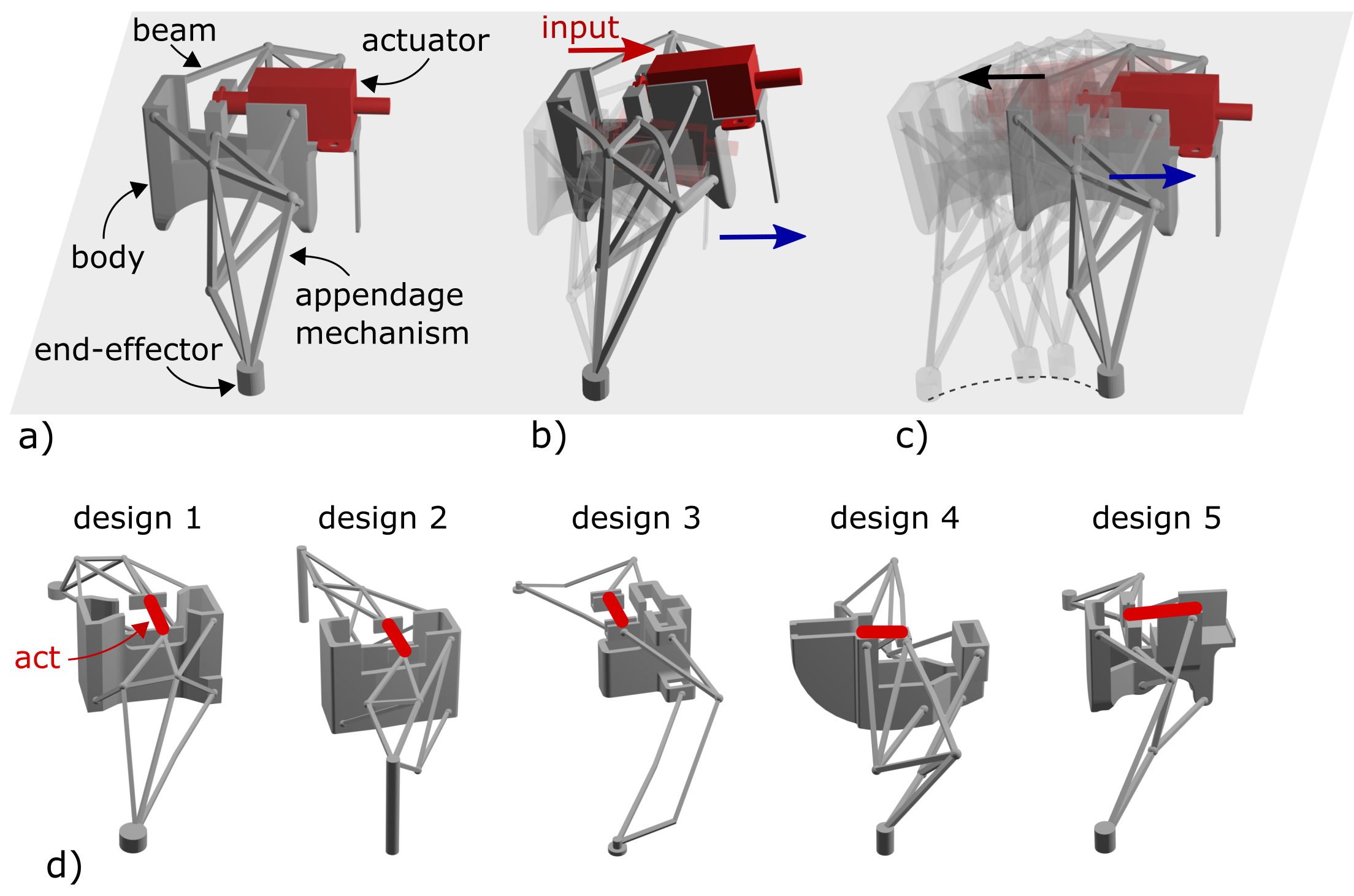}
    \caption[Concept figure]{Concept of soft locomotion robots based on a net of thin elastic beams (spatial compliant mechanisms): (a) illustration of one robot design, (b) working principle - active state, c) working principle – passive state, (d) examples of different robot designs and actuator placements.}
    \label{fig:concept}
\end{figure}

Expanding research interests in the soft robotic field have brought new understanding and knowledge of how soft and elastic materials can be utilized to realize simple yet highly functional soft robots \cite{Jumet2022,Laschi2016}. Unlike rigid-body-based robots composed of rigid links and joints, soft robots are usually built from soft, elastic, or hyperelastic materials \cite{Rich2018}. Traditional rigid body-based systems and robots are often limited in degrees of freedom, maneuvers, and complexity of motions they can achieve, while not being safe to operate around humans and often complicated and expensive to realize. Being inherently soft and continuously deformable, soft robots can overcome these limitations and drawbacks of classical robotic systems. This led to several innovations and applications of soft robots in medicine \cite{Kim2019}, wearable devices \cite{Kim2022,Benjamin2020}, remote exploration and inspection \cite{Shepherd2011,Rich2018}, and food industry and agriculture \cite{Wang2020}.

Different solutions of soft robots are introduced over time: continuous robots in form of manipulators \cite{Fu2020,Asselmeier2020,AlIbadi2018}, robotic grippers for object manipulation \cite{Hao2020, Subramaniam2020, Elfferich2022, Guo2018}, for realizing terrestrial locomotion like walking \cite{Fishman2017, Jiang2020, He2019, Felton2014}, crawling \cite{Pasquier2019, Shepherd2011}, running/galloping \cite{Tang2020}, jumping \cite{Gorissen2020}, multimodal locomotion gaits \cite{Milana2020, Duduta2020, Park2019, Peng2020, Tawk2018}, swimming \cite{Chen2018, Aubin2019}, or underwater exploration \cite{Chu2012}. 
Most of these soft robotic concepts are realized by utilizing pneumatically driven soft elastomer actuators \cite{Shepherd2011, Tawk2018, Gorissen2020, Zolfagharian2021} and combination of these actuators with other elements \cite{Tang2020, Singh2020, Sui2019, Hubbard2021, Santoso2019}. 
Active smart materials \cite{Meng2019, Duduta2020, Guo2021, Park2019, Peng2020, He2019, Shin2018, Ng2020, Tan2021, Tian2020}, tendon driven actuation \cite{Jiang2020, Renda2020, Kakehashi2020}, origami-based structures \cite{Felton2014, Zhai2020, Kim2021, Xu2021, Zou2021}, and tensegrity structures \cite{Lee2020, Rieffel2018} are applied as well for the actuation and development of soft robots. 

Robots based on pneumatically driven soft elastomer actuators are largely realized by utilizing hyperelastic materials \cite{Shepherd2011, Corucci2018, Tang2020, Jumet2022}. Smart material-based robots use other types of active and responsive materials (when exposed to external influences, e.g. heat, light, current) \cite{Meng2019, Duduta2020, Guo2021, Park2019, Peng2020, He2019, Shin2018, Ng2020, Tan2021, Tian2020}, where the material properties often imply slow actuation speed. Origami structures are built from rigid or soft, thin material sheets that can be easily folded \cite{Felton2014, Zhai2020, Kim2021, Xu2021, Zou2021}, where actuation speeds remain relatively slow. Tensegrity soft robots are realized either via a set of rigid rods and springs \cite{Rieffel2018} or from completely soft hyperelastic materials \cite{Lee2020}. Other soft robotic concepts mostly utilize a combination of hyperelastic materials and rigid elements \cite{Rich2018, Santoso2019}. Hyperelastic materials exhibit poor capabilities of rapidly storing and releasing relatively large deformation energy, limiting the possibility to achieve robots with high locomotion speeds. 

Although these concepts lead to the realization of highly functional soft robotic systems, still there are great limitations in actuation speed and payload capacity.

This is mainly due to the mentioned limitations of hyperelastic materials and slow activation time of active smart materials.

Some attempts to increase the soft robot locomotion speed include leveraging the mechanical instability inspired by spine flexion and extension in quadrupedal mammals, such as in the galloping cheetah \cite{Tang2020}. This concept remains dependent on external air supply making onboard actuation hard, while the capability of such robots to realize transport of certain payloads was never proven or tested. 

The design of soft robots in general represents a challenging task. Past work mostly includes developing soft robots based on designer experiences \cite{Jumet2022, Laschi2016, Rich2018} or seeking inspiration from nature, such as from animals \cite{Tang2020, Park2019} or insects \cite{Jiang2020}. 
Methods for automated synthesis of soft locomotion robots remain limited and are mostly focused on pneumatically actuated soft robots \cite{Corucci2018, Cheney2014b, Cheney2014a, Shah2021}, and only a few of these approaches are verified with real-world experiments \cite{Pinskier2022}.

Material sets like common plastic in form of a net of thin elastic beams have rarely been explored for the realization of soft robots \cite{Chen2018, Jumet2022}, however these have the capability to both rapidly store and release deformation energy while providing more strength compared to other mentioned material sets.

Thus, in this paper, we present a different approach to the realization of soft robots formed as a set of spatially distributed thin elastic beams.

The structure of the introduced soft robots leverage compliant mechanisms that are rarely utilized and explored for the realization of soft robots \cite{Chen2018, Vogtmann2011, Pierre2017}. 

Unlike rigid-body based mechanisms comprising of joints and linkages, compliant mechanisms represent monolithic structures that utilize elastic deformation of its individual segments to realize motion and force transmission. Being inherently monolithic, compliant mechanisms offer several benefits over classical ones: reduced complexity, ease of manufacturing, no need for assembly, zero backlash, friction-free motion, no wear, no need for lubrication, better scalability,
better accuracy, lightweight design, built-in restoring force, low cost. 

Due to their many advantages, compliant mechanisms offer a promising solution for innovating in the soft robotics field. Thus, the main goal of this paper is to introduce a different approach to the realization of soft robots, formed as net of spatially connected thin elastic beams and built by using common plastic material. Inspired by compliant mechanism structures, we name this class of robots \emph{Combots} (compliant mechanism robots). As one example of Combots, we present in this paper robots that can realize locomotion. 
The basic concept is illustrated in Fig.~\ref{fig:concept}. The operating principle is based on the transmission of displacement/forces from the input port to the appendage end-effector via elastic deformation of the individual beam-like thin segments (Fig.~\ref{fig:concept}a). Having contact with the ground, the appendages will push the robot body forward, realizing the locomotion of the whole soft robot (Fig.~\ref{fig:concept}b).
With the aim to realize high locomotion speeds while having relatively large available actuation forces in a compact design, we use electromagnetic actuation, in form of linear solenoids, as actuators for the proposed soft robots. By supplying the appropriate electrical power to the solenoid, the plunge will contract and provide the input displacement and force to the robot (Fig.~\ref{fig:concept}b). After the supplied power is turned off, the stored elastic deformation energy in the structure of the appendages together with energy stored in the solenoid spring will be released rapidly, providing in some portion an additional push forward momentum, contributing to the overall robot locomotion, Fig.~\ref{fig:concept}c. We leverage these effects to realize soft robotic systems that can achieve high locomotion speeds compared to other existing soft robots \cite{Tang2020}.

In general, the structure of the Combots resembles spatial compliant mechanisms with distributed compliance. The research of 3D compliant mechanism design methods remains very limited \cite{Patiballa2020, Patiballa2018, Patiballa2019, Krishnan2019, Frecker1997}.
Most soft locomotion robots based on compliant mechanisms are realized by utilizing a planar (2D) compliant mechanism with concentrated compliance \cite{Henning2021}, where compliance is localized across the mechanism structure, in combination with classical motors to actuate the robot locomotion \cite{Vogtmann2011, Pierre2017}. 
This is mainly due to the relatively simple synthesis process, where compliant mechanisms are designed based on existing rigid body mechanisms \cite{Henning2021, Pavlovic2009}. 
Spatial (3D) compliant mechanisms with distributed compliance, where compliance is uniformly distributed across the structure, are not utilized before for the realization of soft locomotion robots, and their potential remains unexplored. This is mainly due to lack of design synthesis approaches and a way to effectively actuate them. 

\begin{figure*}[t]
    \centering
    \includegraphics[width=0.7\textwidth]{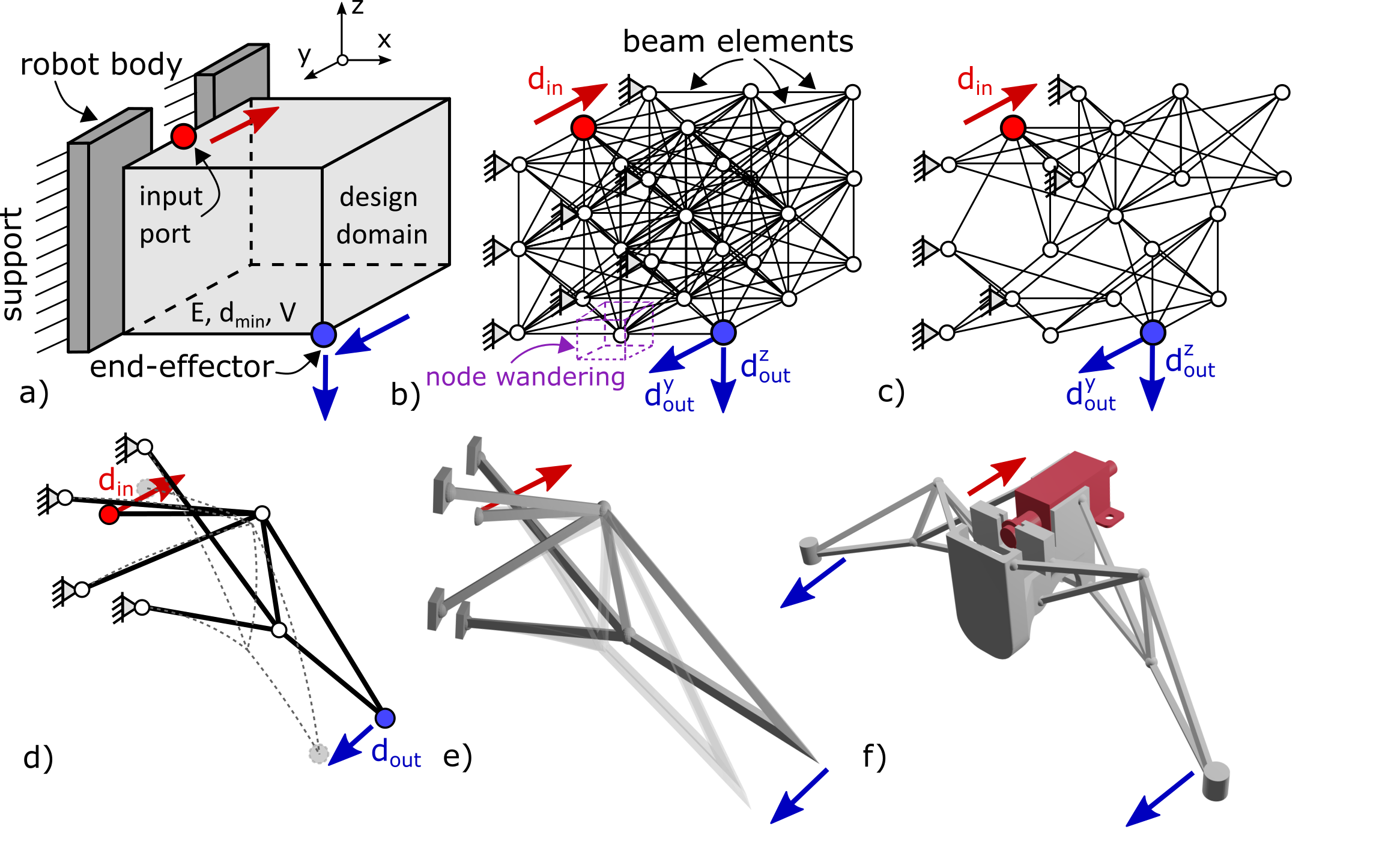}
    \caption[Synthesis framework overview]{Synthesis framework overview for soft locomotion robots: (a) problem formulation, (b) discretization, (c) evaluation of one solution for the optimization process, (d) best found solution, (e) nonlinear Finite Element Method verification, (f) two-leg robot prototype.}
    \label{fig:framework}
\end{figure*}

The many possible design solutions of the proposed soft compliant mechanism robots (Fig.~\ref{fig:concept}d) motivate us to develop an evolutionary algorithm-based synthesis framework for the automated design of Combots. 
This gives a convenient approach to synthesis as no prior design experience is needed to yield an optimal solution, while allowing to explore design solutions beyond those imagined by a human designer. The focus is on the design of one robotic appendage, as based on this two-leg or multiple-leg robots could be developed. 

In addition to proposing a synthesis framework with focus on easy transfer to real-world solutions, 
we demonstrate that the Combots can achieve high locomotion speeds not reported before (up to 3.7 BL/s), with modest strength, while also being capable of transporting a certain payload. 
Moreover, we show that the robots can realize both terrestrial and underwater locomotion. 
This could open up for new application areas, like exploration and inspection in different environments, where such robots are lightweight, easy to transport, require little energy to operate, could be made disposable, and are simple to realize.

The remainder of the paper is organized as follows: 
Section \ref{sec:EA} describes the methodology of the synthesis framework and presents results from the synthesis process. 
Then, section \ref{sec:FEM} presents methods and results from a verification stage applying nonlinear FEM simulations to selected designs. 
Further, section \ref{sec:prototype} describes the prototype realization of the selected designs, and presents results from deformation behavior and locomotion tests. 
Finally, sections \ref{sec:discussion} and \ref{sec:conclusion} discuss the results and conclude the paper.

\section{Related work in design methods}

A structural topology optimization approach is adopted as the basis for the synthesis, mainly derived from the structural mechanics \cite{bendsoe_topology_2004} and compliant mechanisms research fields \cite{Sigmund1997}. 
Topology optimization methods are rarely applied for the design of soft locomotion robots \cite{Caasenbrood2020, Souza2020, Hiller2012}.
In topology optimization, the designer only needs to define the desired inputs (Fig.~\ref{fig:framework}a), while through an evolutionary search algorithm, solutions for the given problem are automatically obtained (Fig.~\ref{fig:framework}f). The typical structural optimization process includes the following steps: topology optimization (finding the optimal material distribution within the given design space), dimensional synthesis (dimensions of individual segments are optimized), and shape optimization (shape of individual structural segments or overall structure is synthesized). For the design of Combot appendages, we focused on the topology optimization process as a more creative part of the synthesis steps including shape optimization, to broaden the possible design search.

The Combot appendages resemble a spatial 3D compliant mechanism structure. Different topology optimization methods for compliant mechanism synthesis are introduced over time \cite{Zhu2020, Sigmund1997, Frecker1997, Kumar2020, Kumar2021a}. But most of these methods focus only on the design of planar 2D compliant structures. Methods for optimal synthesis of spatial 3D compliant mechanisms are rarely introduced as they impose a new level of complexity into the design process, due to spatial distribution of elements and thus relatively large design space \cite{Frecker1997}. 
Recent work \cite{Patiballa2020, Patiballa2018, Krishnan2019} propose a load flow method for the design of spatial compliant mechanisms with distributed compliance, utilizing predefined building block designs to yield a mechanism topology. But such a method is not suitable for the realization of complex appendages, as many of the possible design solutions are disregarded at the beginning and remain unexplored. Other works in the field of spatial compliant mechanism design include using continuum synthesis approaches \cite{Kumar2021, Ansola2010, Huang2014}. These methods are not suitable for the design of discrete beam-like Combot appendages, as they usually lead to structures with concentrated compliance, or solutions that are too stiff for the level of functionality that is required from soft robots. 
Moreover, existing research of spatial compliant mechanisms rarely leads to the fabrication of physical prototypes, where the suitable material solutions remain largely unexplored.

Outside of the compliant mechanisms field, in our past work \cite{Milojevic2022}, we developed a synthesis method for the design of MEMS micro-robot appendages. But this method is still limited to 2D planar-looking topologies and cannot be applied in the case of soft locomotion robots. Thus, a new approach to synthesis needs to be developed.

In general, none of the existing synthesis methods can address the complex design requirements imposed by the Combot appendages. To broaden the possible design search, we incorporated both topology and shape optimization in one synthesis process, which is not presented before for spatial compliant structures. Further, none of the introduced design methods have addressed the problem of synthesis of functional compliant robotic appendages with stroke and payload requirements. Thus, novel objective functions need to be investigated and formulated.

\section{Evolutionary algorithm-based leg mechanism synthesis\label{sec:EA}} 
The synthesis goal for the Combots is to obtain an appendage design that can realize a controllable motion of its end-effector (tip of the robotic leg) in the desired directions when input displacement is applied. Additionally, maximizing the motion transmission (ratio of realizing output to input displacement) and force transmission (ratio of realized output force to applied input force) of the appendages. This is done to realize soft compliant locomotion robots that can utilize actuators with smaller input strokes, while also being energy efficient and capable of carrying a certain payload. The synthesis goal formulation is motivated by the assumption that an appendage end-effector having more contact with the ground, due to higher motion transmission, can lead to Combots that realize high locomotion speeds. In other words, it is a hypothesis that robots with high values of motion transmission ratio can lead to higher locomotion speeds of the overall robot structure while having a modest force transmission ratio, leading to energy-efficient robots.

\subsection{Synthesis framework overview}

\begin{figure*}
    \centering
    \includegraphics[width=0.8\textwidth]{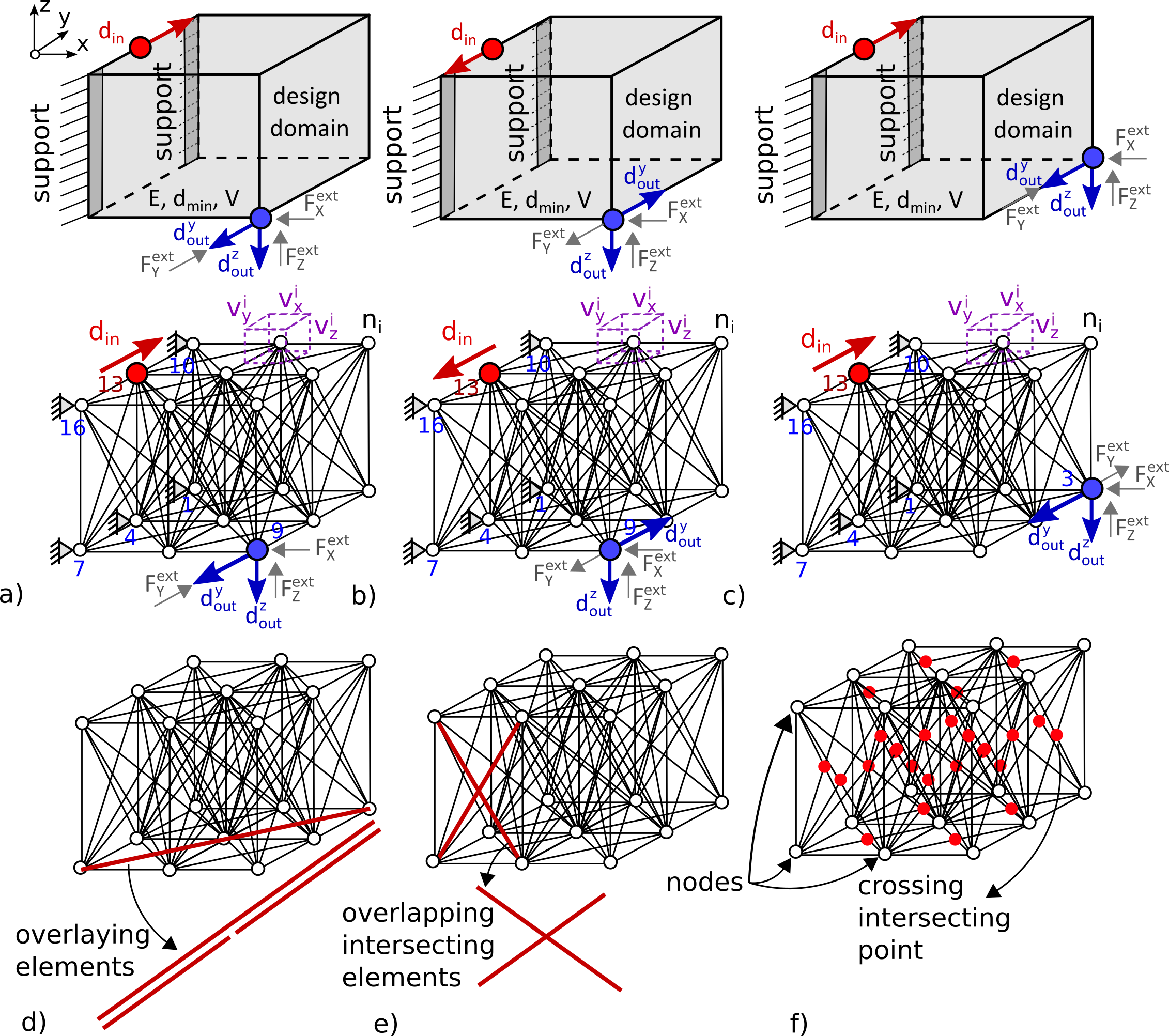}
    \caption[Problem formulation]{(a-c) Problem formulation and discretization of the design domain for the soft compliant locomotion robot synthesis. Different investigated cases, where input and output displacement direction are varied together with location of the appendage end-effector point. (d) Fully connected ground structure with overlaying elements, (e) partially connected ground structure containing overlapping intersecting elements, (f) ground structure with identified crossing - intersecting points.}
    \label{fig:formulation}
\end{figure*}

The proposed synthesis method is illustrated in Fig.~\ref{fig:framework}. The goal is to optimize the design of one robotic appendage, as two- or multiple-leg soft robots could be developed based on the same appendage design.
The Combot appendages consist of a net of spatially connected thin beam-like segments, where connections between elements are modeled as point connections. This simplifies the analysis and is an advantage of the synthesis method.
The main synthesis steps include problem formulation (Fig.~\ref{fig:framework}a), discretization (parameterization) (Fig.~\ref{fig:framework}b), and optimization (Fig.~\ref{fig:framework}c). 

The problem formulation consists of defining the design domain shape and 
boundaries,
input displacement/force of the actuator, 
location of the input port where the actuator is attached, location of the end-effector, desired end-effector displacement direction, location of supports (where the appendage is attached to robot body), external loads that act on the appendage (to simulate the resistance of the ground), material characteristics,
and other constraints (like the desired density of the solution, or minimum required output displacement). 

As a next step, the discretization of the design domain is realized (Fig.~\ref{fig:framework}b). Due to the discrete nature of the beam-like design of the appendages, the Ground Structure Approach (GSA) is utilized for the parameterization \cite{Milojevic2022}, here expanded for the spatial problems. 
The design domain is divided by a number of nodes and a set of beam elements spatially connecting the nodes representing the ground structure. Other methods for parameterizing include block-based methods \cite{Grossard2009} or continuum discretization \cite{Zhu2020}, but these are not applicable for the case of spatial Combot appendages synthesis, due to limited exploration of the design space and different problem formulation.
The discretized design domain i.e. network of elements shown in Fig.~\ref{fig:framework}b represents the initial solution from which the Combot appendage design is searched. In the proposed synthesis process, individual elements can be removed from or returned to the ground structure (Fig.~\ref{fig:framework}c). 
Additionally, the overall shape of the appendages is optimized by allowing the individual nodes to wander within a predefined region (Fig.~\ref{fig:framework}b), changing the position and length of the elements.

After the discretization, a search method needs to be applied to find the optimal solution within the given design space. The discrete nature of the problem motivates us to utilize discrete optimization methods, of which Evolutionary algorithms (EAs) are especially suitable, as 
they have proven to be efficient when searching large design spaces and have already been successfully applied in the topology optimization of compliant-based structures \cite{Milojevic2022, Parsons2002}.
An objective function has to be provided for the EA, and in our case two alternative functions are proposed, where one focuses on the calculated geometric advantage, and the other also includes the mechanical advantage of the appendage design.
For the evaluation of the objective function, a linear finite element method analysis is applied.
After the optimization has converged the remaining elements in the structure together with the chosen node locations will form the design of the Combot appendages that can potentially lead to Combot locomotion (Fig.~\ref{fig:framework}d).

Based on the obtained solution a solid 3D model of the appendage is designed (Fig.~\ref{fig:framework}e). To verify the appendage deformation behavior and functionality a nonlinear Finite Element Method simulation is performed (Fig.~\ref{fig:framework}e). As a final step, the appendage structure is fabricated via a fuse deposition modeling process by using a 3D printer, after which the linear solenoid actuator is attached (Fig.~\ref{fig:framework}f).

\subsection{Problem formulation}

The main goal is to design a Combot appendage that can realize the output motion of its end-effector in the desired direction when input displacement is applied, leading to overall robot locomotion. Here for the problem formulation, the cuboid design domain is defined (Fig.~\ref{fig:formulation}a). Fixed supports are placed at the left edge areas of the design domain, representing places where the appendage will be attached to the robot body. Input port (where the actuator will be attached) is located at the middle of the left upper edge of the design domain. As an input, a displacement is applied (actuator available stroke), where the direction of input motion is defined in a Y direction (it is assumed that the used actuator has enough available input force to actuate the appendages). The direction of the input displacement ultimately depends on how the actuator will be positioned, allowing different design options to be explored. The end-effector point is located at the design domain right side, placed at the down edge lower point. The desired direction of output motion is set in opposite from the direction of the input displacement, and in the -Z direction (Fig.~\ref{fig:formulation}a-c). This is motivated by the aim to realize robotic appendage that can both move the robot body in the desired direction of motion (here Y direction) while also pushing the end-effector against the ground to lift the robot body from the ground and realize overall robot locomotion (motion in -Z direction). The external forces are applied at the end-effector point (in all three directions opposite the desired output displacement direction) to simulate the resisting force of the ground. Additionally, to this, a material characteristic (Young modulus) from which the appendages will be produced is defined as well (here the whole robotic appending is produced from a single material, but multiple materials can be used). The specific parameter values that are used for the problem formulation are given in Table~\ref{tab:params}.

To demonstrate different design possibilities with the proposed synthesis method, three different cases are investigated (Fig.~\ref{fig:formulation}a-c), where direction of the input and output displacement are varied, together with location of the appendage end-effector point.

\subsection{Parameterization of the design domain}

To realize automated synthesis the problem formulation and design domain needs to be represented by a set of variables that can be optimized. The design domain is discretized by using the ground structure approach (GSA) \cite{Milojevic2022, Parsons2002}, but here developed for spatial problems (Fig.~\ref{fig:formulation}a-c). The cuboid design domain is divided by $n_x \times n_y \times n_z$ number of nodes and a network of spatial beam elements connecting these nodes. For the element formulation, spatial beam elements with 6 degrees of freedom at each end are used. The ground structure in Fig.~\ref{fig:formulation}a-c represents the initial solution or design space within the optimal solution is searched for. The design variables are:
\begin{itemize}

\item $v^i_{el}$ - defines if the selected element is present or absent from the ground structure.
  The variable can take discrete values $ v^i_{el} \in \{ 0, 1 \} $; 
  denoting absence and presence in the structure, respectively.
 The total number of $v^i_{el}$ variables is equal to the total number of elements, $i = num_{elem}$.

\item $v_x^j, v_y^j, v_z^j$ – defines the position of the nodes. 
The robotic appendage shape is optimized by allowing the individual nodes to wander within the predefined spatial region, changing the lengths and position of the elements. 
All the variables can take discrete values within the predefined range (Table~\ref{tab:params}). 
The number of position variables 
is equal to the total number of the nodes in the structure, $j = num_{nodes}$. 
\end{itemize}

Initially, all the nodes in the structure are interconnected with one beam element, leading to a fully connected ground structure (Fig.~\ref{fig:formulation}d). 
Such a structure contains overlaying elements 
that are difficult to produce (Fig.~\ref{fig:formulation}d). 
To avoid obtaining such solutions a filtering algorithm
is applied that eliminates overlaying elements from the structure prior to the optimization. 

The ground structure can also contain overlapping elements, intersecting each other at one or multiple positions (Fig.~\ref{fig:formulation}e). 
The appendages with such elements are difficult to manufacture, and lead to stiffer robotic solutions. Thus, an algorithm is developed that identifies the total number of crossing points in the structure (Fig.~\ref{fig:formulation}f), where this number is then minimized during the optimization. 
The problem of eliminating such elements has not been solved before in the topology optimization of spatial compliant structures.

Depending on the grid resolution, the total number of variables can vary (e.g. for the design case in Fig.~\ref{fig:formulation}a-c, the overall number of variables is 232). Table~\ref{tab:params} contains the rest of the design domain parameterization parameters.

\subsection{Objective formulation for optimization}

The initial ground structure represents a large design space containing a possible solution to the set problem. Search method in form of optimization is applied to find the optimal appendage design. The main goal of the optimization is to obtain Combot appendages that can realize output displacement of the end-effector in desired directions when input displacement is applied. This is motivated by the end aim to realize Combot designs that can achieve locomotion in the desired direction while selecting the direction of input actuation. Two different synthesis cases are investigated here, thus two main objectives are defined:
\begin{itemize}
\item Maximizing the motion transmission ratio of the appendage, that is, the ratio of realized output displacement of the end-effector to the applied input displacement. This is defined as the \emph{Geometric Advantage}, $GA =  d_{out} / d_{in}$.

\item Maximizing the force transmission ratio of the appendage, that is, the ratio of realized output force at the end-effector point to the needed input force to deform the appendage. This is defined as the \emph{Mechanical Advantage}, $MA = F_{out} / F_{in}$.
\end{itemize}

These are done to explore the trade-offs between Combot performances, designed when only GA is optimized and when both GA and MA are taken into consideration. Here the hypothesis is that appendages, where only GA is considered, would realize higher output displacement of the end-effector while requiring larger input forces to actuate the robot, where Combot design when both GA and MA criterias are optimized, would be more energy efficient but at the cost of realizing smaller output displacement. These fundamental trade-offs leave space for optimization, with a goal to realize robotic appendages that can achieve both high GA and MA values, and potentially higher locomotion speeds.

In addition, three more objectives are introduced to the optimization process:
\begin{itemize}
    \item To maximize the appendages payload capacity or yield a solution that is stiff enough to realize locomotion and carry a payload, the end-effector displacement due to external forces, $d_{out}^{ext}$, is minimized. 
    \item To prevent obtaining solutions with non-existing or very small densities, the desired density of the structure of the appendages is optimized. The optimization goal is implemented by minimizing the difference between the realized 
    ( $L^{rel}_{tot} = \sum_{i=1}^{n_{rel}} L_i $)
    and the desired ($L_{des}^{tot}$) sum of all element lengths in the structure. 
    \item The number of overlapping, or crossing, elements are eliminated by minimizing the total number of identified overlapping points, $n_{overlap}$, in the given solution.

\end{itemize} 

The automated synthesis of the Combot appendages represents a multi-objective optimization problem. We follow a weighted sum approach to reduce the objectives into a single objective function for the evolutionary algorithm.

As two synthesis cases are investigated, different formulations of the objective function are defined.

The objective function when only GA is considered, in the optimization, is given as follows:
\begin{equation}
\label{eqn:ga}
max[GA - w_1 \cdot d^{ext}_{out} - w_2 \cdot |L_{rel}^{tot} - L_{des}^{tot}| - w_3 \cdot n_{overlap}]
\end{equation}

The objective function when both GA and MA are considered is given in the following form:

\begin{equation}
    \label{eqn:gama}
    max[GA - w_1 \cdot d^{ext}_{out} - w_2 \cdot |L_{rel}^{tot} - L_{des}^{tot}| - w_3 \cdot n_{overlap} - w_4 \cdot \frac{1}{MA}]
\end{equation}

The weights used for the objective functions can be found in Table~\ref{tab:ea_params}, and are varied as described in Sec.~\ref{sec:synth_results}.
The terms in the objective functions are calculated by using linear analysis. 

For this synthesis case the initial problem setup is modified by adding a spring element at the end-effector tip to calculate the work done by 
 the agent on the external environment. 
The realized output force is calculated as force required to deform the spring element attached at the end-effector tip.

\subsection{Evolutionary algorithm}
\begin{table}
    \caption{Evolutionary Algorithm parameters.}
    \label{tab:ea_params}
    \centering
    \begin{tabular}{|l|c|}
    \hline
    Parameter & Value \\
    \hline
    population size & 200 - 400 \\
    number of generations & 1000 \\
    selection function & roulette \\
    crossover probability & 95\% \\
    mutation probability & 9\% \\
    elite count & 2 \\
    \hline
    $w_1$ & 0.3 : 0.1 : 0.7 * $10^2$  \\
    $w_2$ & 0.3 : 0.1 : 0.5 * 10 \\
    $w_3$ & 1 \\
    $w_4$ & 0.1 : 0.1 : 0.6 \\
    \hline
    \end{tabular}
\end{table}
  
The design of Combot appendages represents a discrete optimization problem, as elements are either present or absent from the initial structure, while also other variables have discrete values. 
This motivates us to use an Evolutionary Algorithm (EA), in particular the \emph{Genetic Algorithm}\footnote{We will refer to the genetic algorithm as the EA to avoid confusion with the geometric advantage (GA) to be optimized.} variant, as these algorithms are considered to be robust in discrete, complex, and relatively large search spaces.
Genetic algorithms have previously shown good performance in structural optimization~\cite{goldberg1986engineering,rajeev1992discrete,rajan1995sizing,deb2001design}, and have also proven to be efficient for solving different optimization problems in the compliant mechanisms field~\cite{Parsons2002, Milojevic2022}.

The EA works on a population of candidate solutions, which are coded as bit strings. The length of a single solution varies with the specific parameterization of the optimization case to be carried out, such as the total number of nodes and elements, outlined in Table~\ref{tab:params}.
Selection is performed using the roulette wheel method, one-point crossover and bit flip mutations are used, and a generational replacement method with elitism is employed. 
If variation operators lead to solutions which do not satisfy the constraints, these are discarded and the operations are applied again until valid solutions are found.

Details on the EA parameters can be found in Table~\ref{tab:ea_params}.

\subsection{Synthesis results\label{sec:synth_results}}
\begin{figure*}
    \centering
    \includegraphics[width=0.8\textwidth]{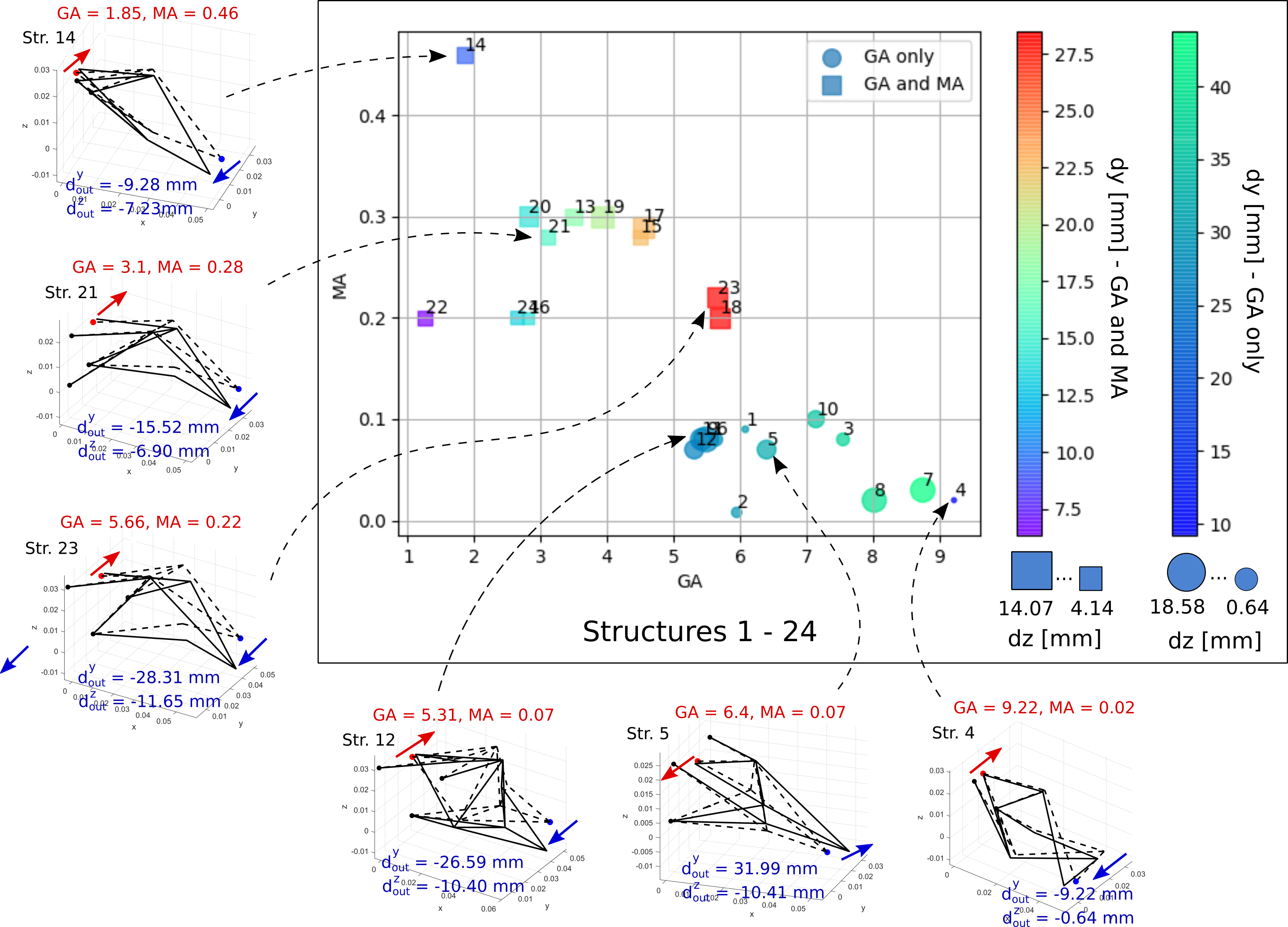}
    \caption[Results syntheses]{Results of the soft compliant appendage robot synthesis. Solutions where only motion transmission ratio is optimized (GA values) are marked with circle. Solutions where both motion and force transmission ratio are optimized (MA values) are marked with rectangle. Results are shown for the various problem setups defined in Fig.~\ref{fig:formulation}a-c. Solutions 1-4 correspond to problem setup at Fig.~\ref{fig:formulation}a, Solutions 5-8 correspond to problem setup at Fig.~\ref{fig:formulation}b, Solutions 9-12 correspond to problem setup at Fig.~\ref{fig:formulation}c. This is repeated for solutions 13-24, where both GA and MA were optimized for the same setups.
    Circles signify structures optimized for GA only, while squares signify optimization for both GA and MA.
    }
    \label{fig:synth_results}
\end{figure*}

To obtain Combot appendages solutions multiple Genetic Algorithms optimizations are run for each of the set synthesis cases (varying weighting constants and initial problem settings,  Table~\ref{tab:ea_params}, Table~\ref{tab:params}). The optimization simultaneously eliminates elements from the initial structure (or allows them to reappear), while moving the structure nodes within the predefined region. The remaining elements with selected positions of the nodes will form the optimal solution of the soft compliant mechanism robot appendage. 
Synthesis results for both design cases are summarized in a combined plot in Fig.~\ref{fig:synth_results}, together with example structures. 
The following subsections discuss the results for the investigated design cases.

\subsubsection{Optimizing geometric advantage only}

The circles in Fig.~\ref{fig:synth_results} show the performance of the solutions when motion transmission ratio criteria are optimized as described with Eq.~\ref{eqn:ga}. Different appendage designs are obtained for various weighting constants and initial problem parameters definition, as shown in Fig.~\ref{fig:formulation}a-c (to explore different design trends, see Table ~\ref{tab:params}). Examples of these designs can be seen in the bottom part of Fig.~\ref{fig:synth_results}.
The results demonstrate that in all cases the soft robotic appendages realize motion of the end-effector in the desired direction when input displacement is applied.

The appendage's design performances are compared considering GA, MA, and realized displacement values in the Y and Z direction of the end-effector point (Fig.~\ref{fig:synth_results}), as these have the greatest effect on the robot locomotion. Taking performance parameters into account, Structure 7 and 8 are of special interest. 
As can be seen, these designs can realize relatively high GA values and realize large displacements in Y and Z directions, but at the cost of having low MA values, thus requiring a relatively large input force to actuate them.  
The optimization also managed to find some solutions where there is a trade-off between GA and MA, Structures 5 and 10, while realizing modest values of end-effector displacement in both Y and Z directions. Solutions in this region represent the area of interest with highest probability to realize locomotion. 

\subsubsection{Optimizing the mechanical and geometric advantage}

The squares in Fig.~\ref{fig:synth_results} show the performance of the solutions when both mechanical and geometric advantage criteria are taken into consideration, following Eq.~\ref{eqn:gama}.
In all cases optimization led to appendage designs that can realize the desired direction of the end-effector motion.
Here the special focus is on solutions Structure 13, 19 - 21. 
These appendage designs realize both relatively high MA values while having modest GA values, with a modest range of end-effector displacement in desired directions. 
Not surprisingly Structure 14 realizes a high MA but at the cost of lower GA values, and Structures 18 and 23 realize high GA but at the cost of lower MA values. 
In general, the EA manages to find designs where there are good trade-offs between GA and MA, such as Structures 19-21. 
It is hypothesized that these solutions have a higher probability to realize locomotion while also being capable of carrying a payload. 
In general, it could be concluded that these designs are more energy-efficient, while it is left to verify if they can lead to better locomotion performance.

To further study locmotion performances, appendage designs 5 (only GA optimized) and 21 (GA and MA optimized) are selected, as different representative solutions of each investigated case.

\section{Nonlinear FEM deformation behavior investigations\label{sec:FEM}}
To investigate the deformation behavior of the obtained Combot appendages solutions, a nonlinear finite element method analysis is performed by using commercially available FEM software (Abaqus). The investigations are done for the appendage design solutions that were selected as representative designs of the studied optimization cases (Structure 5 and 21, Fig.~\ref{fig:synth_results}).
The solutions were chosen to represent different optimization cases and different objective functions (GA vs. GA and MA).

\subsection{Method}
Based on the obtained solutions, a solid 3D model of the appendages is realized (Fig.~\ref{fig:FEM}a, d). For the FEM simulations, a fixed boundary condition is applied at the surfaces where appendages should be attached to the robot body. 
To simulate the actuator stroke, displacement of d\textsubscript{in} = 5 mm (same as in the optimization) is applied at the input port surface.
For the solver, a large deformation analyst setup is defined. Fig.~\ref{fig:FEM} shows the complete initial simulation setup (a, d) as well as the mesh model of one appendage (b, e). As different synthesis cases are investigated (Fig.~\ref{fig:formulation} and ~\ref{fig:synth_results}), FEM simulations are run for each of the investigated cases.

\subsection{Results}
\begin{figure*}
    \centering
    \includegraphics[width=0.8\textwidth]{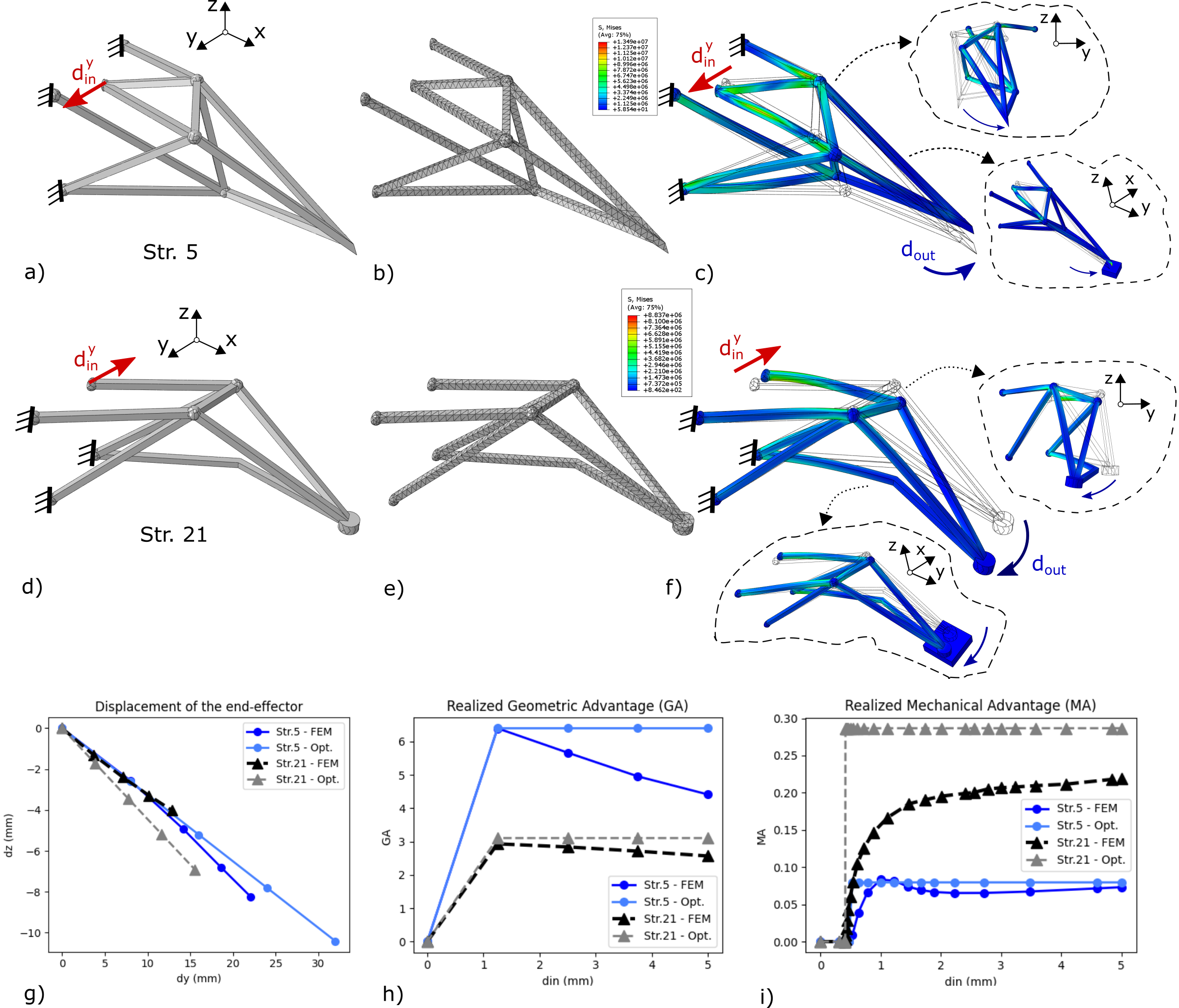}
    \caption[FEM]{Nonlinear finite element method (FEM) analyses of appendage deformation behavior: FEM simulation setup (a, d), mesh model of the appendages (b, e), deformation behavior (c, f). Results of FEM investigations: g) end-effector output displacement, h) realized Geometric Advantage, i) realized Mechanical Advantage.}
    \label{fig:FEM}
\end{figure*}

Fig.~\ref{fig:FEM}c shows the appendage's deformation behavior for the case when only GA was optimized, while Fig.~\ref{fig:FEM}f when both GA and MA were optimized. In all cases, simulation results show that similar deformation behavior is realized as in the solutions obtained with optimization 
(Structures 5 and 21 in Fig.~\ref{fig:synth_results}). 
The end-effector realizes motion in the desired direction when input displacement is applied.

To investigate the appendage motion path, the realized end-effector displacement in the Y and Z directions is plotted (Fig.~\ref{fig:FEM}g). The achieved values of GA (Fig.~\ref{fig:FEM}h) and MA (Fig.~\ref{fig:FEM}i) with respect to applied input displacement, are analyzed as well. All the results are compared with values obtained from optimization. For smaller values of end-effector displacement, there is a good agreement between optimization and FEM results (Fig.~\ref{fig:FEM}g), structures behave linearly.

As the output displacement increases, nonlinear deformation effects are more expressed, thus there is a difference between the results (linear analysis is used in the optimization). This is more evident in the case of Str. 21, while in general in the case of Str. 5 there is a better agreement between the results (Fig.~\ref{fig:FEM}g). Similar trends could be observed when investigating the realized GA (Fig.~\ref{fig:FEM}h) and MA values (Fig.~\ref{fig:FEM}i). When in the linear regime of deformation behavior there is a good agreement between results, while for a higher input displacement, there is a difference in results (for GA, Fig.~\ref{fig:FEM}h, better agreement in case of Str. 21, while for MA, Fig.~\ref{fig:FEM}i, better agreement in case of Str. 5i).

Considering the performance parameters (displacement, GA, and MA values), Str. 5 realizes a larger output displacement compared to Str. 21 (Fig.~\ref{fig:FEM}g). This is also reflected in the GA of the appendages, where Str. 5 archives higher GA values compared to Str. 21 (Fig.~\ref{fig:FEM}h). Not surprisingly, Str. 21 realizes higher MA values compared to Str. 5, leading to a more energy-efficient solution (smaller input forces are required to deform the appendage structure). In general, Str. 21 realizes a better trade-off between GA and MA values.

\section{Prototype realization and locomotion experiments\label{sec:prototype}}

Two Combot designs are selected for realization as representative results from each of the investigated synthesis cases (structures 5 and 21). In this paper, the focus is on realizing two-legged robots, but three, four, or multiple leg robots could be achieved as well. Based on the obtained solutions (Fig.~\ref{fig:synth_results}) and FEM investigations (Fig.~\ref{fig:FEM}a, d), 3D solid models of the two-leg soft compliant mechanism robots are designed (Fig.~\ref{fig:prototype}a). The following subsections describe a fabrication process and prototyping of the two-leg Combots. Further, the experimental investigations of robots’ deformation behavior, characteristics, and locomotion capabilities under various conditions are presented.

\subsection{Prototyping and fabrication}
\begin{figure*}
    \centering
    \includegraphics[width=0.8\textwidth]{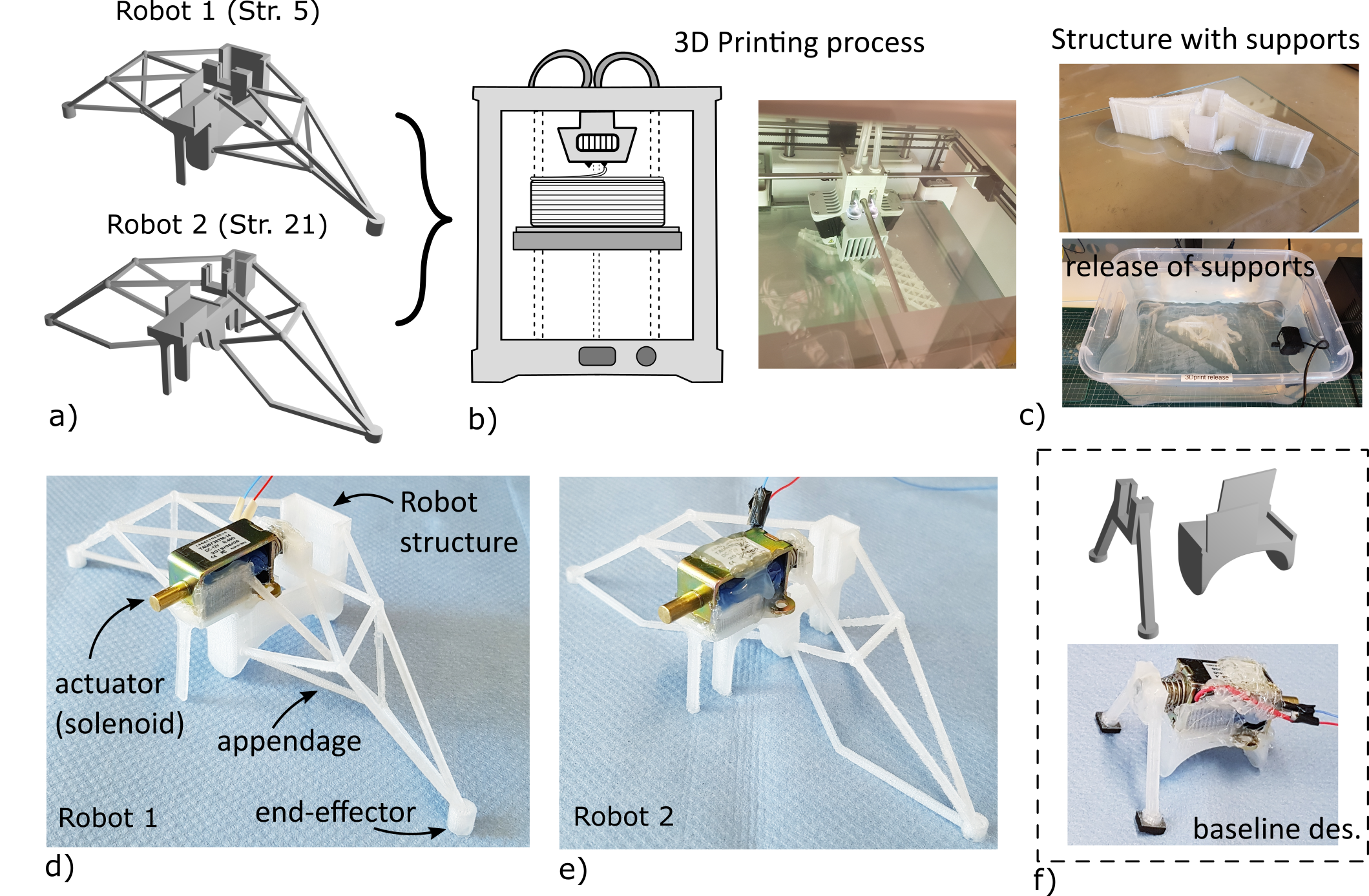}
    \caption[Prototyping]{Prototyping and fabrication of the \emph{Combot} soft compliant mechanism robots: a) two-legged soft robot designs, b) fused deposition modeling process – 3D prating, c) produced structure with supports and realize of the supports, d) final produced robot prototype based on solution Str.5 – Robot 1, e) produced prototype based on the appendage solution Str. 21 – Robot 2, f) baseline design and produced robot prototype.}
    \label{fig:prototype}
\end{figure*}

The robot fabrication is realized via the fuse deposition modeling process by utilizing a commercially available 3D printer Ultimaker (Fig.~\ref{fig:prototype}b). Substantial research efforts are made to determine the suitable material and printing process parameters for the Combot prototypes. Several filament materials are explored: ABS, Ultimaker PP, BASF PP, and different settings for the printing process (values of the material Young modulus are given in Tab.~\ref{tab:materials}). As Combot designs represent a complex structure comprised of spatially connected beam-like elements, the supports need to be used to yield the production of a firm robot structure. The supporting structure geometry is automatically generated via Ultimaker Cura software, while the PVA (polyvinyl alcohol) soluble material filament is used for the support fabrication (Fig.~\ref{fig:prototype}c). Thus, the multi-extrusion 3D printing is realized with two materials, one for the Combot structure and the other for the supports. Depending on the complexity and density of the Combot structure, the 3D printing process time took on average 20 hours. After the production, the soft robots are submerged in a water-based solution, to release the main Combot structure from the support material (Fig.~\ref{fig:prototype}c). 

Based on multiple trial-and-error approach experimentation with various materials, fabrication of different robot designs, and considering the needed actuation forces to deform the Combot structure, it was determined that Ultimaker PP filament led to the best results.
 
This takes into consideration robot functionality and printing quality: small actuation forces are needed to achieve deformation, while having a structure firm enough to realize robot locomotion. 
Thus, PP filament was adopted as the final material for the soft compliant mechanism robot’s fabrication. Fig.~\ref{fig:prototype} shows the 3D printing process (Fig.~\ref{fig:prototype}b), produced structure with supports, the release of the support material (Fig.~\ref{fig:prototype}c), and final prototypes of the two selected Combot designs (Fig.~\ref{fig:prototype}d, e).

\begin{figure*}
    \centering
    \includegraphics[width=0.8\textwidth]{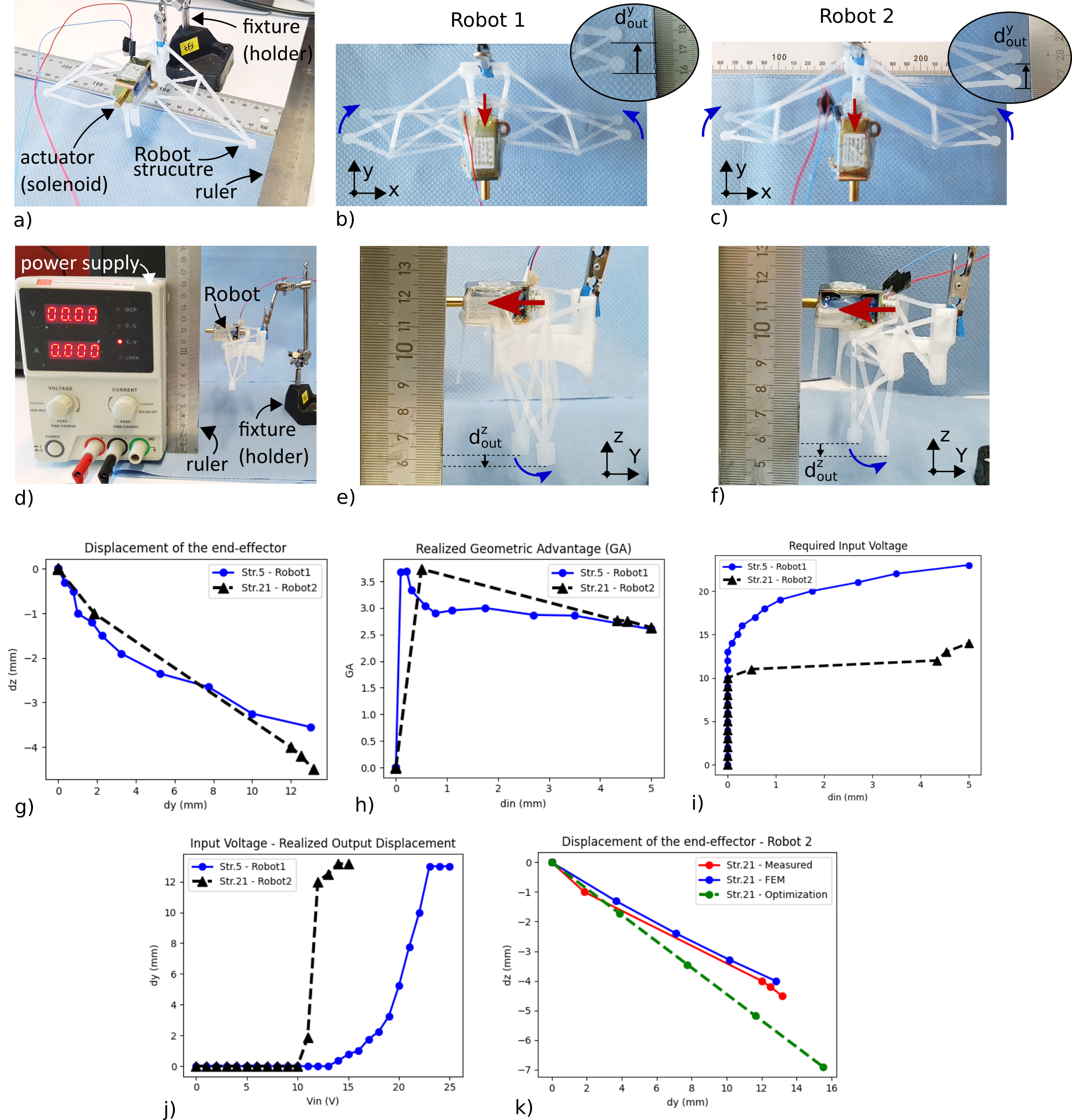}
    \caption[Mesurments deformation]{Experimental investigation of soft compliant mechanisms robot deformation behavior. Setup for measurement of end-effector displacement in the Y direction (a). Robot 1 (b) and Robot 2 (c) are shown in initial and deformed positions (X-Y plane). Setup for measuring output displacement in the Z direction (d). Robot 1 (e) and Robot 2 (f) are shown in initial and deformed positions (Z-Y plane). Measurement results for different investigated cases (g-k).}
    \label{fig:measurements_exp}
\end{figure*}

To realize autonomous actuation of the Combots, an appropriate actuator is needed. Motivated by the aim to achieve high locomotion speeds of the Combots, while having relatively large available actuation forces (in compact design) with modes power consumption, we adopted the electromagnetic actuation principle in form of a common solenoid. The actuation in form of solenoids is rarely applied or used in the soft robotics field, thus offering new insights into how these actuators can be utilized to drive soft robots. The solenoids offer advantages of realizing relatively large actuation forces and high actuation speeds (by controlling the solenoid frequency) while requiring reality low actuation power. Some drawbacks are a poor ratio of actuator overall size to possible actuator stroke while being relatively heavy compared to Combot structure. The solenoid with 5 mm available actuation stroke is attached to the soft compliant mechanism robot structure (at predefined dedicate space), where the solenoid plunger is connected to the robot input port. The actuator position is secured by fixing (gluing) the solenoid body to the Combot body structure. Fig.~\ref{fig:prototype}d, e shows the final prototypes of soft compliant mechanism robots with integrated actuation.

Additionally, to the optimized designs in Fig.~\ref{fig:prototype}a, a simple intuitively developed solution (baseline design) is produced as well (Fig.~\ref{fig:prototype}f). This is motivated by the aim to prove the advantages of the proposed synthesis approach and Combot locomotion capabilities when compared to a simple design solution with solenoid actuation. This is also done to demonstrate that Combot locomotion does not come from the solenoid actuation/vibrations only and that appendages mechanism design has a large influence on the overall robot motion capabilities. The baseline design is realized in form of two-beam elements and a robot body where the solenoid is attached, Fig.~\ref{fig:prototype}f, the same filament material is used as in the case of Combot fabrication. In such a design, a motion transmission ratio is GA = 1, where the MA$\approx$1, meaning that there is no amplification of the applied input displacement. The hypothesis is that low values of GA will lead to slow locomotion speeds of the overall robot proving the benefits of realizing soft compliant mechanism robots through the proposed synthesis approach.

\subsection{Deformation behavior}
\begin{figure*}
    \centering
    \includegraphics[width=0.9\textwidth]{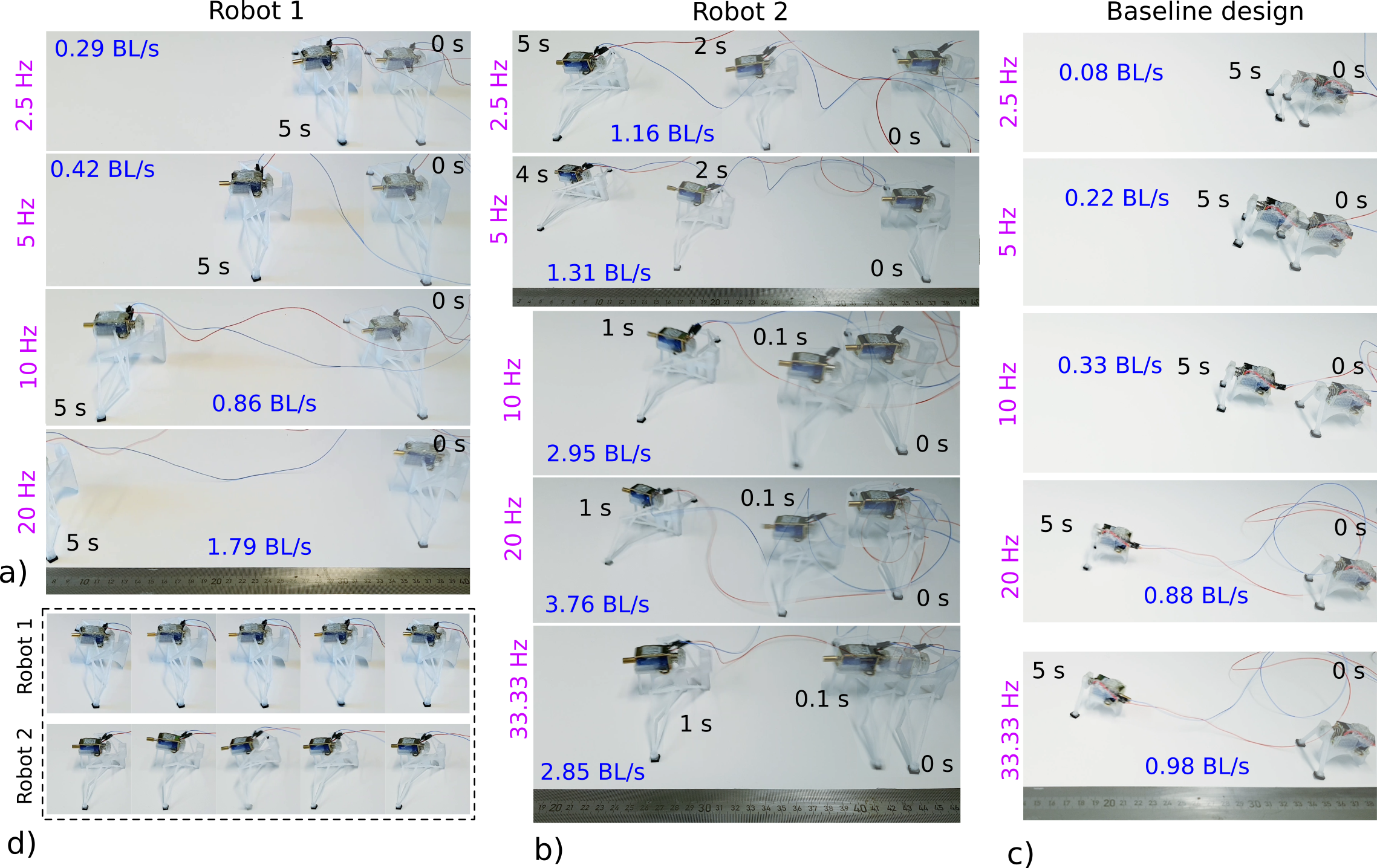}
    \caption[Locomotion speed]{Experimental investigation of soft compliant mechanism robots locomotion capabilities and speed: (a) Robot design 1, (b) Robot design 2, (c) baseline design (robot position is shown in different time intervals and at different actuation frequencies), (d) comparison of locomotion principles of robot~1 and robot~2. See \href{https://youtu.be/TEjqIhG39N0}{video 1}.}
    \label{fig:speed_locomotion}
\end{figure*}

The measurement setup for experimental investigation of the Combot deformation behavior (output motion of the appendage end-effector) is shown in Fig.~\ref{fig:measurements_exp}a, d. The robot body is fixed above the ground to enable the free movement of the appendages, while two rulers are placed to visually represent the values of applied input and realized output displacement of the appendage end-effector Fig.~\ref{fig:measurements_exp}a (in case of measuring output displacement in the Z direction, one ruler is used Fig.~\ref{fig:measurements_exp}d). Additionally, the value of applied input voltage for the solenoid is recorded. The input voltage is supplied to the solenoid in increments of 1 V (in range of 0 – 25 V for Robot 1, and 0 – 15 V for Robot 2), while in each step the solenoid stroke (input displacement) and end-effector displacement is measured. Computer Vision software is used to track the trajectory of the appendage end-effector point while also allowing to realize accurate displacement measurements. 

The same measurement procedure is applied for all analyzed Combot designs (Fig.~\ref{fig:measurements_exp}b, c, e, and f). Fig.~\ref{fig:measurements_exp}g shows results of end-effector output displacement in Y and Z direction (similar to Fig.~\ref{fig:FEM}g), and Fig.~\ref{fig:measurements_exp}h shows the realized GA values compared to applied input displacement, for both Combot designs. Further, the correlation between solenoid input stroke and required input voltage (Fig.~\ref{fig:measurements_exp}i), and the correlation between the input voltage and realize output displacement in the Y direction (Fig.~\ref{fig:measurements_exp}j) are plotted. As could be seen (Fig.~\ref{fig:measurements_exp}h), in general, GA values are dropping with input displacement increment, where both Robots 1 and 2 realize a similar range of end-effector displacement (Fig.~\ref{fig:measurements_exp}g). Surprisingly Robot 2 archives a larger value of displacement in both Y and Z directions. For intermediate values of actuator stroke, higher values of GA are realized in the case of Robot 2 (Fig.~\ref{fig:measurements_exp}h). When full actuator stroke is reached, a similar range of GA values is realized in the case of both Robot 1 and Robot 2 (Fig.~\ref{fig:measurements_exp}h). This is contrary to FEM simulation results (Fig.~\ref{fig:FEM}h). In the case of Robot 1, results are not in good agreement with simulation, due to the manufacturing imperfections of the robot design and connection problems between the solenoid and robot input port. Fig.~\ref{fig:measurements_exp}k shows the comparison between results obtained with optimization, FEM simulation, and measurements, for Robot 2. As could be seen there is a good agreement between FEM and measured displacement of the end-effector point. 

Based on the results in Fig.~\ref{fig:measurements_exp}i, j, it could be concluded that a larger input voltage is required to actuate Robot 1 (based on Str. 5) compared to Robot 2 (based on Str. 21). These results agree with trends observed in Fig.~\ref{fig:FEM}i, where Str. 21 achieve better MA values compared to Str. 5 (meaning it requires smaller input forces to actuate/deform the structure). This proves that the synthesis case when both GA and MA are considered in the optimization leads to more energy-efficient Combot solutions, that also can realize relatively high GA values. Although there is a difference when results are compared to measurements, overall, the optimization can lead to appendages solutions that realize similar deformation behavior when the solutions are experimentally tested (both Combot designs realize end-effector displacement in the desired direction as the set goal of the optimization).

\subsection{Terrestrial locomotion}
\begin{figure}
    \centering
    \includegraphics[width=0.47\textwidth]{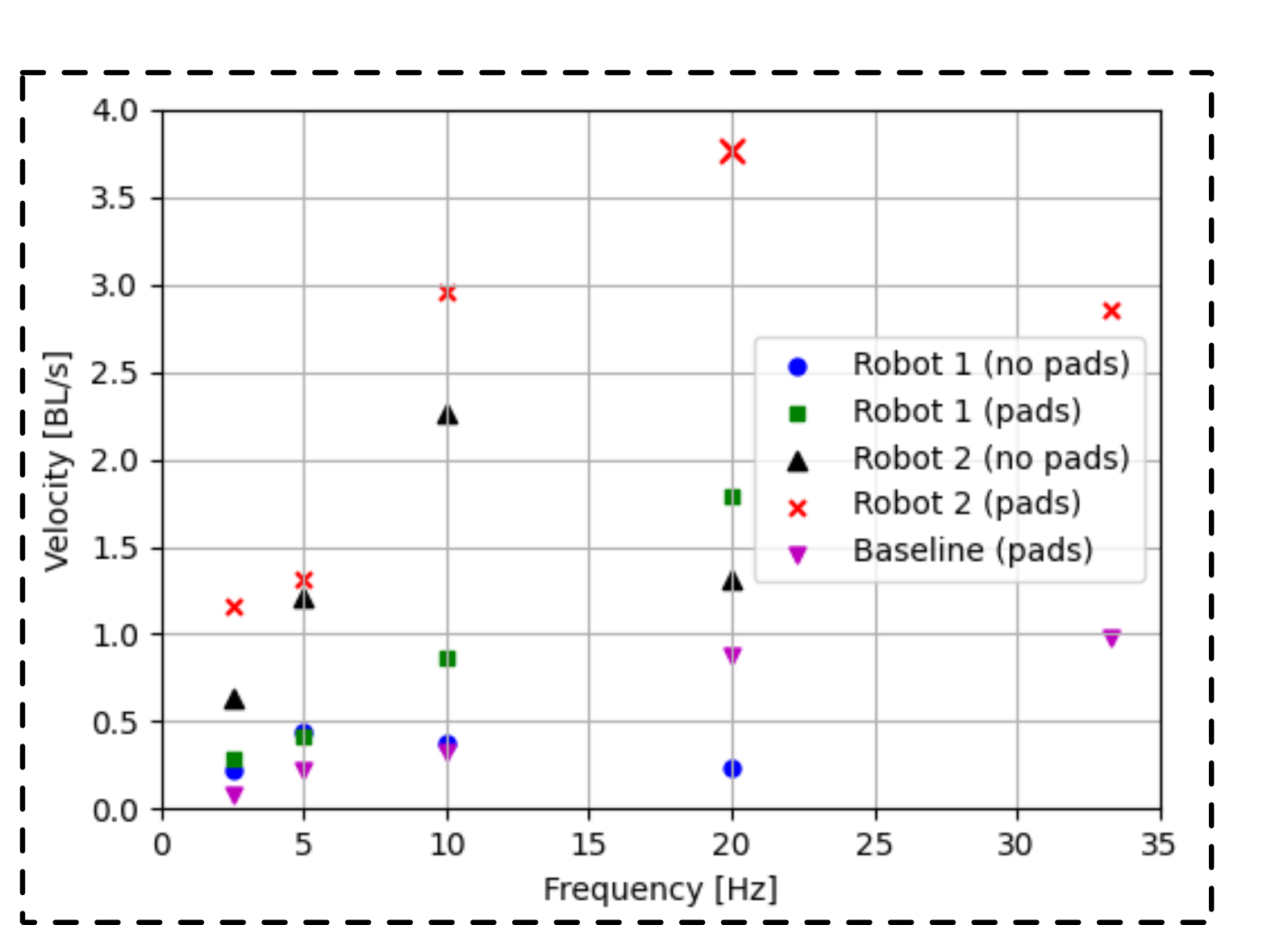}
    \caption[Locomotion speed plot]{Comparison of locomotion speeds at different control frequencies for the different robots shown in Fig.~\ref{fig:speed_locomotion}.}
    \label{fig:speed_locomotion_plot}
\end{figure}

The Combot designs (Fig.~\ref{fig:prototype}d, e) locomotion characteristics and capabilities are investigated by realizing motion on the flat smooth surface. The ruler is used to visually represent the robot’s locomotion speed (Fig.~\ref{fig:speed_locomotion}). Here, several experiments are done:
\begin{itemize}
    \item actuating the solenoid manually by pressing a switch, where a constant voltage is supplied to the actuator,
    \item actuating the solenoid via a controller where a constant voltage is supplied, and the actuation frequency is varied (in range of 2.5 to 33 Hz),
    \item for both cases (manual actuation and using the controller), the Combot locomotion is investigated without and with adding anti-slip pads to the end-effectors to increase the friction between the ground and robots.
\end{itemize}
The same experiments are done for all Combot designs (Fig.~\ref{fig:prototype}d, e). Fig.~\ref{fig:steps_locmotion}, Fig.~\ref{fig:speed_locomotion}d shows the robot operation principle i.e. locomotion principle captured in different consecutive time steps. Two motion phases could be observed, which we will refer to as the \emph{active} and the \emph{passive} phases:

In the first active phase, when the voltage is supplied to the actuator the solenoid input stroke is transmitted to the end-effector via appendages mechanism structure. The appendages push the robot (with the solenoid) agents to the ground (Fig.~\ref{fig:steps_locmotion}), realizing the lift and forward motion of the robot body (due to designed end-effector free motion in Y and -Z direction, Fig.~\ref{fig:measurements_exp}c, f). In the second passive phase, when the solenoid is turned off, the stored deformation energy of the appendage’s mechanism, together with the elastic energy of the deformed solenoid spring, is released. This provides, in some portion, additional push forward momentum (work) enabling the whole Combot structure to detach from the ground and realize forward jump-like motion (Fig.~\ref{fig:steps_locmotion}, Fig.~\ref{fig:speed_locomotion}d) (considering that returning path of the end-effector is revers of the graph in Fig.~\ref{fig:measurements_exp}g, thus not having contact with the ground). By leveraging these two effects/phases, high locomotion soft robot speeds can be achieved.

Realized Combot locomotion (all robot designs) for different investigated cases and at different actuation frequencies, is shown in Fig.~\ref{fig:speed_locomotion} (captured in different time intervals). Fig.~\ref{fig:speed_locomotion} shows results when pads are added to the robot end-effectors to increase the friction with the ground, but the same investigations are realized for robots without pads (Fig.~\ref{fig:steps_locmotion}).

Achieved robot locomotion speeds are plotted in Fig.~\ref{fig:speed_locomotion_plot}.

Based on the results, several trends could be observed. Not surprisingly, higher locomotion speeds are realized at higher actuation/solenoid frequencies, for most Combot designs (Fig.~\ref{fig:speed_locomotion_plot}). In the case of Robot 2 (Fig.~\ref{fig:speed_locomotion}b), it is noted that at frequencies above 30Hz, the locomotion becomes unstable, as the solenoid induces high vibrations into the system, having phases where appendage end-effectors do not realize contact with the ground, thus leading to drops in locomotion speed. Still, relatively high velocities are achieved. 

Comparing results with and without an anti-slip pad, in general robots achieve higher speeds when the pads are added to the robot, due to increase friction with the ground (Fig.~\ref{fig:speed_locomotion_plot}). Interestingly, Combot design 2 can achieve high locomotion speeds up to 3.76 BL / s (body length per second, where body length is measured in direction of robot motion), at an actuation frequency of 20Hz, outperforming some of the existing soft locomotion robots (even at the lower solenoid frequencies of 10Hz) \cite{Tang2020}. Additionally considering that robots are carrying relativity heavy solenoids. Combot design 2 (Fig.~\ref{fig:speed_locomotion}b), realizes higher locomotion speeds compared to Combot design 1 (Fig.~\ref{fig:speed_locomotion}a). Even when comparing Robot 2 with no pads and Robot 1 with pads (Fig.~\ref{fig:speed_locomotion_plot}). This is because in Robot design 2 both GA and MA are optimized (based on Str. 21 in Fig.~\ref{fig:synth_results}, ~\ref{fig:FEM}d), leading to a more energy-efficient solution, as more of maximal available solenoid force is at Combot disposal (in the case of Combot design 1, a maximal possible solenoid actuation force is needed to drive the robot). Additionally, to this, due to higher MA values in the case of Robot 2, the transmitter output force at the appendages end-effector is larger, leading to pushing the robot body with larger forces. Moreover, results show that even relatively lower values of GA can lead to faster soft robots if MA is optimized i.e. there is a good trade-off between GA and MA values. 

Compared to baseline design (Fig.~\ref{fig:speed_locomotion}c) Robot designs 1 and 2 (Fig.~\ref{fig:speed_locomotion}a, b) realize higher locomotion speeds (Fig.~\ref{fig:speed_locomotion_plot}). 

\subsection{Locomotion with payload}
\begin{figure*}[t]
    \centering
    \includegraphics[width=0.90\textwidth]{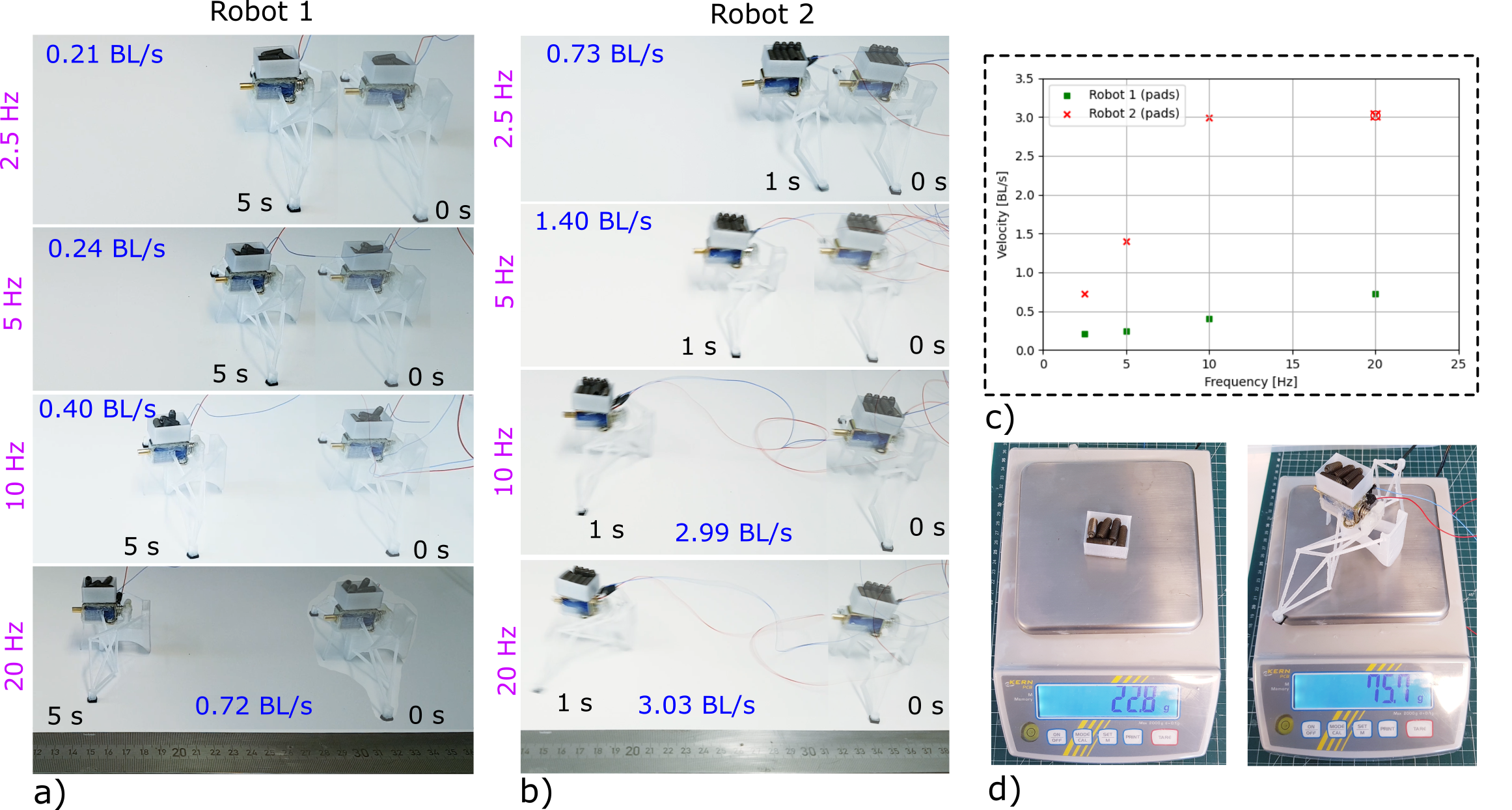}
    \caption[Locomotion payload]{Experimental investigation of robot locomotion when carrying a payload: (a) Robot design 1, (b) Robot design 2, (c) comparing robot locomotion speed, (d) weight of the payload and overall robot. See \href{https://youtu.be/CILk9G_DMws}{video 2}}
    \label{fig:payload_locomotion}
\end{figure*}

To prove that introduced concept of soft compliant mechanism robots can also carry a payload while realizing locomotion (Fig.~\ref{fig:payload_locomotion}), similar investigations are done like in the case of (Fig.~\ref{fig:speed_locomotion}). A container in the shape of a box is 3D printed and filled with small items (bolts), the payload values are shown in Fig.~\ref{fig:payload_locomotion}d. The experiments are realized by placing/attaching the container on top of the Combot solenoid, where the robot locomotion capabilities are then tested at different actuation frequencies (Fig.~\ref{fig:payload_locomotion}a, b). These investigations are done for all Combot designs. Fig.~\ref{fig:payload_locomotion}a, b shows soft robots realizing locomotion, captured in different time intervals, while the realized locomotion speeds are plotted in Fig.~\ref{fig:payload_locomotion}c. Based on the results it could be concluded that in all cases robots can realize stable locomotion and carry a payload. Not surprisingly smaller values of locomotion speeds are achieved (Fig.~\ref{fig:payload_locomotion}c), with similar trends like in the case of Combots without the additional payload (Fig.~\ref{fig:speed_locomotion_plot}). 

In the case of Robot design 2, high locomotion speeds up to 3.03 BL/s could be realized (Fig.~\ref{fig:payload_locomotion}b).

\subsection{Underwater environment locomotion}
\begin{figure*}[t]
    \centering
    \includegraphics[width=0.90\textwidth]{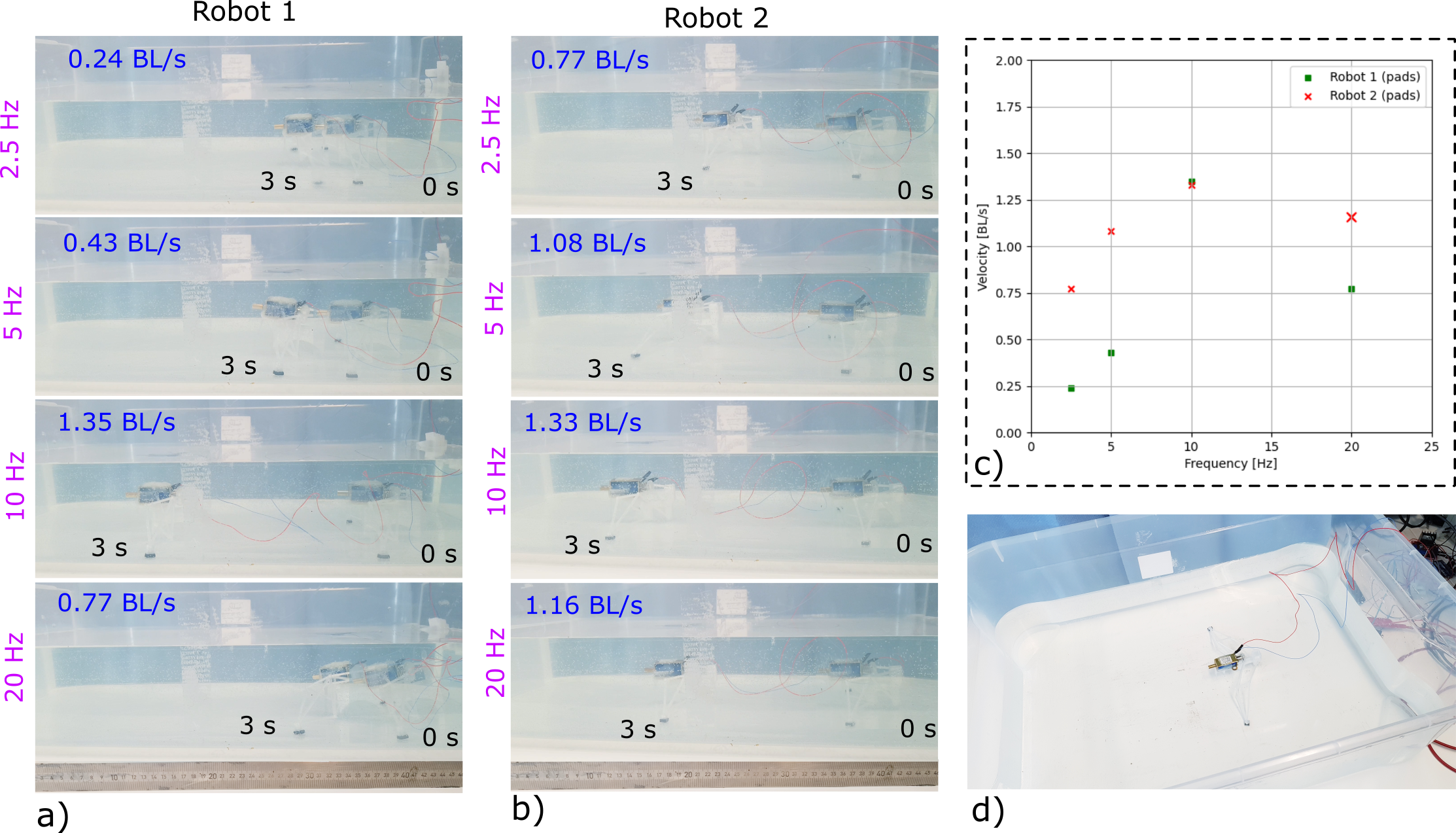}
    \caption[Underwater]{Experimental investigation of robot locomotion underwater (robot location is shown in a different time interval and at various actuation frequencies): (a) Robot design 1, (b) Robot design 2, (c) comparison between robot achieved locomotion speed, (d) plain view of Robot design 2 in the water container. See \href{https://youtu.be/g-g07rmkWUM}{video 3}}
    \label{fig:underwater}
\end{figure*}

The inherent monolithic structure of the soft compliant mechanism robots and type of used actuation, allow Combots to operate in other environments besides terrestrial. To experimentally investigate this, robots are fully submerged underwater (Fig.~\ref{fig:underwater}). Similar investigations are realized like in the case of terrestrial locomotion (Fig.~\ref{fig:speed_locomotion}, ~\ref{fig:payload_locomotion}). Combots are actuated via the controller by varying the actuation frequency (Fig.~\ref{fig:underwater}). The same investigations are done for both robot designs. Fig.~\ref{fig:underwater} shows Combot locomotion capabilities realized underwater (captured in different time intervals). The achieved locomotion speeds are plotted in Fig.~\ref{fig:underwater}c. Similar deformation behavior is exhibited like in the case of terrestrial locomotion, but with several additional effects. Due to the different environments, robots realize more jumping-like motion, with several phases of the robot body being detached fully from the ground when robots are actuated. Still, the robot body is heavy enough to bring the Combot to the ground. 
In general, results show that the introduced concept of soft compliant mechanism robots can realize stable locomotion even underwater. Not surprisingly, the locomotion speeds are lower than in the case of terrestrial operation (Fig.~\ref{fig:speed_locomotion_plot}). Robot design 2 (on average) realizes higher locomotion speeds compared to Robot design 1, reflecting similar trends like in (Fig.~\ref{fig:speed_locomotion_plot}, ~\ref{fig:payload_locomotion}c). At actuation frequencies of 20Hz, both robot designs realize slower locomotion, due to solenoid inducing higher vibrations into the system thus Combots lose contact with the ground and realize unstable motion. 

\section{Discussion\label{sec:discussion}}
Overall, from our experiments with the synthesis framework, the FEM validation, the prototyping, and the locomotion experiments, we observe the following:
\begin{itemize}
    \item \emph{Synthesis approach.}
    Overall, the EA design approach allowed us to explore a range of different designs according to our different specified constraints, and solutions had good performance.
    The EA managed to find solutions that were good tradeoffs when both GA and MA were considered in the objective function, whereas when considering only on GA, the resulting MA performance was relatively low. 
    Potentially, a native multi-objective EA such as NSGA-II \cite{deb2002fast} could produce a better pareto front of design tradeoffs, and should be investigated in future studies. 
    Likewise, while the discovered solutions were relatively diverse, potentially Quality-Diversity EA variants \cite{pugh2016quality} could facilitate a more systematic exploration of various design alternatives.
    
    \item \emph{FEM Validation.}
    We can see that for the solution where GA was optimized to a high value, a higher discrepancy was observed. This is due to the large displacements that led to a large deformation behavior of the appendages. These effects could not be captured by the linear analysis during the optimization process.
    A similar effect could be seen for the MA, for the solution with the highest MA value. 
    Although there is a difference between the results predicted by the optimization process and the nonlinear FEM simulations, similar trends could be observed in both cases, meaning that the optimization could still capture the desired behavior of the appendages.
    This means that the synthesis framework can be executed at practical speeds, avoiding the computationally expensive nonlinear FEM simulations for every candidate solution evaluation.
    
    \item \emph{Prototyping.} 
    Finding the right material for fabrication required substantial trial and error. 
    It was necessary to find good a tradeoff bewtween the required input force and the structural stiffness, which both affect the robot locomotion. 
    For our type of actuator, we ended up with Ultimaker PP as a good tradeoff. For other types of actuation, other materials may have better properties and should be explored.
    
    \item \emph{Real-World Locomotion.} 
    The results from real-world locomotion experiments demonstrated successful transfer of the synthesized solutions, and shoved that the proposed synthesis approach can lead to an optimized soft compliant mechanism solution that realizes better locomotion capabilities than a simple intuitive design. 
    Moreover, it is also clear that the combined optimization of motion (GA) and force (MA) transmission ratios, can lead to new soft robot solutions that realize fast locomotion, outperforming the design considering only GA. 
    When considering other optimization methods, taking into account both GA and MA is likely to lead to better performing solutions.
    
    The proposed concept of soft compliant mechanism locomotion robots is also capable of carrying a certain payload, which is rarely explored for other soft robots \cite{Shepherd2011, Tang2020, He2019, Corucci2018}.
    Moreover, although several concepts of soft robots that can realize locomotion both on the ground and underwater have been introduced \cite{Rich2018, Corucci2018, Chu2012}, the main limitation is slow locomotion. 
    The locomotion speed with payload of 3.03 BL/s even outperformed some of the existing soft locomotion robots \cite{Tang2020}, and underwater speed is also relatively high. This can further broaden the application scope of the Combots.
    
    For different robot structures, payloads, and environments, different actuation control strategies may be optimal. Tuning the shape and frequency of these could be subject of future studies.

\end{itemize}

\section{Conclusion\label{sec:conclusion}}
This paper introduces a different approach to realizing soft locomotion robots --Combots-- by utilizing a net of thin-elastic beams, distributed in form of a spatial compliant mechanism, and fabricated from common PP plastic material on a commercial 3D printer. The inherent material properties as rapid store/release of deformation energy, coupled with electromagnetic actuation in form of a common solenoid, allow us to achieve a higher locomotion speed of soft robots, compared to some existing concepts. 
The spatially connected beam-like elements with the possibility to form different robot topologies allow exploration of a much broader design space compared to other types of soft locomotion robots.

A synthesis framework for obtaining automated design of soft locomotion robots is presented. 
By utilizing the developed systems, different solution designs are obtained, investigated under various problem setups. 

Based on the obtained results, two Combot designs were selected for experimental testing of robot locomotion capabilities. A two-leg soft complaint mechanism robot was designed and fabricated via fused deposition modeling by using a commonly available 3D printer. 

Various experimental investigations are done, testing the robot locomotion capabilities. The results show that both robots could successfully locomote, however the design considering both GA and MA had the best performance for both terrestrial locomotion, locomotion with a payload, and underwater locomotion.

Future work will include developing a systems framework where soft compliant mechanism robots are designed based on interaction with the environment and their locomotion capabilities, realized in a simulated environment. 
We will also consider more advanced optimization methods and different appendage designs that can realize directional locomotion.


\section*{Acknowledgements}
This work was supported by the Academy of Finland Research Council for Natural Sciences and Engineering, grant number 318390, and partially supported by the Research Council of Norway through its Centres of Excellence scheme, project number 262762.



\bibliographystyle{ieeetr}
\bibliography{references,references_soft_robotics}

\appendix
\counterwithin{figure}{section}
\counterwithin{table}{section}
\setcounter{figure}{0}
\setcounter{table}{0}

\subsection{Design parameter space}
Design parameters for the synthesis of the appendages are listed in Table~\ref{tab:params}.
\begin{table*}
\centering
\caption{Design parameters for synthesis of soft compliant robotic appendages.}
\label{tab:params}
\begin{tabular}{|l|c|c|c|} 
\hline
Design parameters & Case 1 & Case 2 & Case 3 \\ \hline
Domain size (length x width x height) & 50 x 30 x 20 mm & 50 x 30 x 20 mm & 50 x 50 x 30 mm \\ \hline
node grid size & \multicolumn{3}{|c|}{$n_x \times n_y \times n_z = 3 \times 3 \times 2$} \\ \hline
degree of nodal connectivity & \multicolumn{3}{|c|}{1} \\ \hline
total number of elements & \multicolumn{3}{|c|}{89} \\ \hline
input node & \multicolumn{3}{|c|}{13} \\ \hline
end-effector node & 9 & 9 & 3 \\ \hline
support nodes & \multicolumn{3}{|c|}{1, 4, 7, 10, 16} \\ \hline
input displacement & \multicolumn{3}{|c|}{$d_{in}$ = 5 mm} \\ \hline
node wandering region size $v_x \times v_y \times v_z $  & \multicolumn{3}{|c|}{$v_x$ = 1 ÷ 2 mm, $v_y$ = 1 ÷ 2 mm, $v_z$ = 1 ÷ 2 mm} \\ \hline
element dimensions (w x h x t) & \multicolumn{3}{|c|}{1 x 1 x 1 mm} \\ \hline
element material (Young modulus) & \multicolumn{3}{|c|}{E = 800 MPa} \\ \hline
external loads& \multicolumn{3}{|c|}{$F_x$, $F_y$, $F_z$ = 1 N} \\ \hline
minimal desired end-effector displacement & \multicolumn{3}{|c|}{$d_{out}^{des}$ = 1 mm} \\ \hline
desired structure density& \multicolumn{3}{|c|}{$L_{tot}^{des}$ = 123 ÷ 230 mm} \\ \hline
\end{tabular}
\end{table*}

\subsection{Material and 3D printing parameters}
The materials used for experimentation and their corresponding 3D printing parameters are listed in Table~\ref{tab:materials}.
\begin{table*}
\centering
\caption{Filament materials and 3D printing parameters for robot fabrication}
\label{tab:materials}
\begin{tabular}{|l|c|c|c|c|}
\hline
Filament materials & Ultimaker ABS & Ultimaker PP & Ultrafuse® PP Natural (BASF) & Ultimaker PVA \\ \hline
Modulus (MPa) & 2070 & 305 & 470 - 554 & - \\ \hline
Nozzle & \multicolumn{4}{|c|}{0.4 AA} \\ \hline
Quality (mm) & 0.06 & 0.08 & 0.08 & 0.08 \\ \hline
Line width (mm) & 0.35 & 0.38 & 0.38 & 0.35 \\ \hline
Infill & 20\% - triangle & 20\% - octet & 20\% - octet & 20\% - triangle \\ \hline
Printing Temp & 225 & 205 & 230 & 215 \\ \hline
Plate temp & 80 & 85 & 75 & 85 \\ \hline
Printing speed (mm/s) & 50 & 25 & 25 & 35 \\ \hline

\end{tabular}
\end{table*}

\subsection{Locomotion steps}
\begin{figure*}
    \centering
    \includegraphics[width=0.8\textwidth]{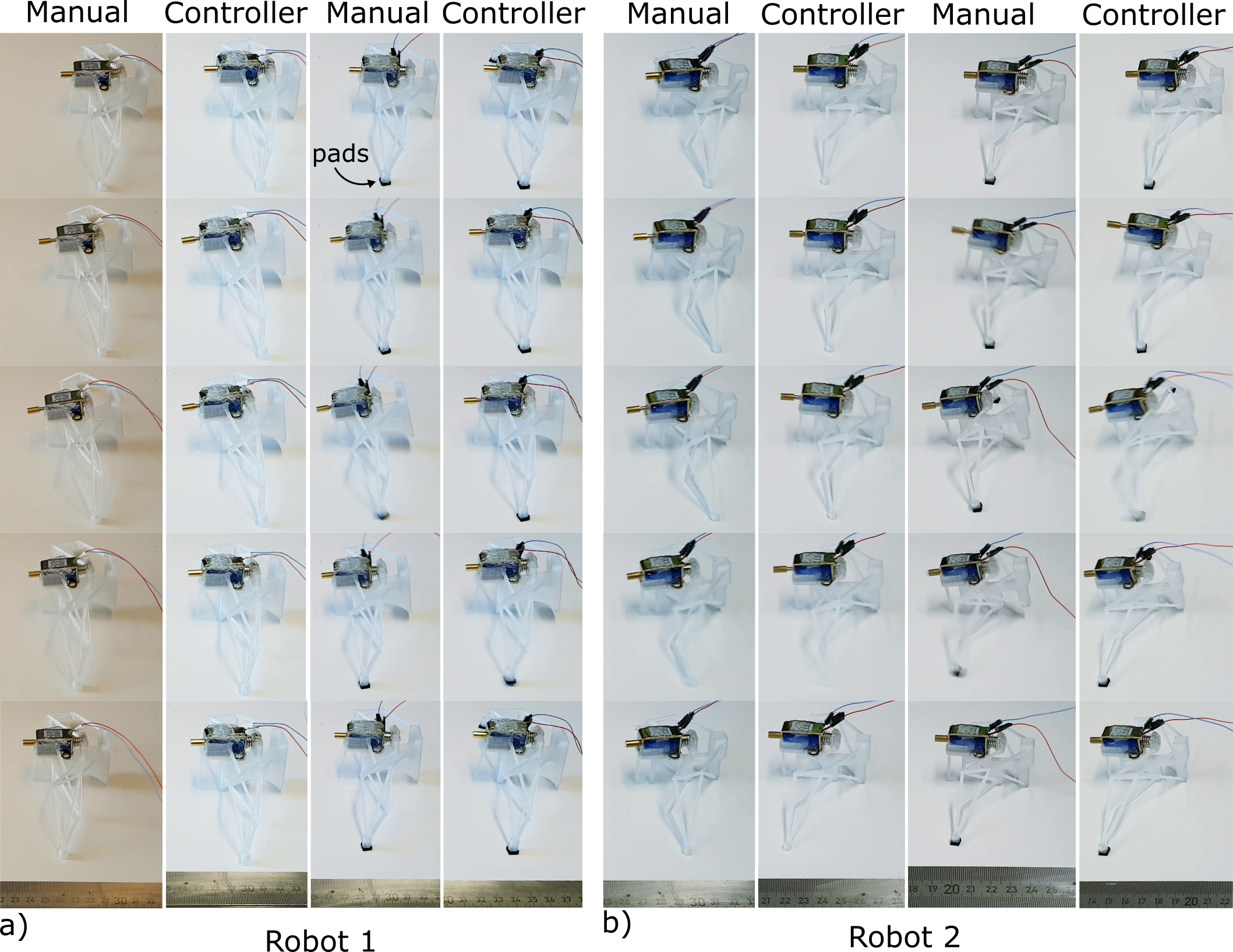}
    \caption[Locomotion steps]{Locomotion principle: (a) Robot 1 design, (b) Robot 2 design. The columns containt key points in the locomotion cycle. Cases are shown for manual control and when using a controller, and without and with pads to increase friction.}
    \label{fig:steps_locmotion}
\end{figure*}

Key frames from videos of the locomotion cycle for the different robots are shown in Fig.~\ref{fig:steps_locmotion}.

\end{document}